\documentclass[AMA,STIX1COL]{WileyNJD-v2}

\articletype{Article Type}%

\received{}
\revised{}
\accepted{}
        
\begin{document}

\title{Autotuning PolyBench Benchmarks with LLVM Clang/Polly Loop Optimization Pragmas Using Bayesian Optimization (extended version)}

\author[1]{Xingfu Wu*}
\author[2]{Michael Kruse}
\author[1]{Prasanna Balaprakash}
\author[2]{Hal Finkel}
\author[1]{Paul Hovland}
\author[1]{Valerie Taylor}
\author[3]{Mary Hall}

\authormark{X. Wu \textsc{et al}}

\address[1]{\orgdiv{Mathematics \& Computer Science Division}, \orgname{Argonne National Laboratory}, \orgaddress{\state{IL}, \country{USA}}}
\address[2]{\orgdiv{Argonne Leadership Computing Facility}, \orgname{Argonne National Laboratory}, \orgaddress{\state{IL}, \country{USA}}}
\address[3]{\orgdiv{Department of Computer Science}, \orgname{University of Utah}, \orgaddress{\state{UT}, \country{USA}}}

\corres{*Xingfu Wu, Argonne National Laboratory, The University of Chicago. \email{xingfu.wu@anl.gov, wuxf@uchicago.edu}}

\presentaddress{Mathematics \& Computer Science Division, Argonne National Laboratory, Lemont, IL 60439, USA}

\abstract[Summary]{In this paper, we develop a ytopt autotuning framework that leverages Bayesian optimization to explore the parameter space search and compare four different supervised learning methods within Bayesian optimization and evaluate their effectiveness. We select six of the most complex  PolyBench benchmarks and apply the newly developed LLVM Clang/Polly loop optimization pragmas to the benchmarks to optimize them. We then use the autotuning framework to optimize the pragma parameters to improve their performance. The experimental results show that our autotuning approach outperforms the other compiling methods to provide the smallest execution time for the benchmarks syr2k, 3mm, heat-3d, lu, and covariance with two large datasets in 200 code evaluations for effectively searching the parameter spaces with up to 170,368 different configurations. We find that the Floyd-Warshall benchmark did not benefit from autotuning because Polly uses heuristics to optimize the benchmark to make it run much slower. To cope with this issue, we provide some compiler option solutions to improve the performance. Then we present loop autotuning without a user's knowledge using a simple mctree autotuning framework to further improve the performance of the Floyd-Warshall benchmark. We also extend the ytopt autotuning framework to tune a deep learning application.}


\maketitle


\section{Introduction}

As the complexity of high performance computing (HPC)  ecosystems (hardware stacks, software stacks, applications) continues to rise, achieving optimal performance becomes a challenge. The number of tunable parameters an HPC user can configure has increased significantly, resulting in the overall parameter space growing dramatically. Exhaustively evaluating all parameter combinations becomes very time-consuming. 
Therefore, autotuning for automatic exploration of parameter space is desirable.
An autotuning is an approach that explores a search space of possible implementations/configurations of a kernel or an application by selecting and evaluating a subset of implementations/configurations on a target platform and/or use models to identify the high performance implementation/configuration within a given computational budget. 
Such a strategy, however, requires search methods to efficiently navigate the large parameter search space of possible configurations in order to avoid expensive training while retraining just enough information to determine optimal configurations or implementations. 
In this work, we develop machine learning (ML)-based autotuning frameworks to reduce the parameter search space in order to autotune loop optimization pragmas to improve performance.

Traditional autotuning methods are built on heuristics that derive from experience \cite{TC02, CH06, GO10} and model-based methods \cite{TH11, BG13, FE17}. At the compiler level, machine-learning-based methods are used for automatic tuning of the iterative compilation process \cite{OP17} and tuning of compiler-generated code \cite{TC09, MS14}. Recent work on autotuning of OpenMP code has gone beyond loop schedules to look at parallel tasks and function inlining \cite{SJ19, KC14, MA11}. In particular, a lightweight framework was proposed to enable autotuning of OpenMP pragmas to ease the performance tuning of OpenMP codes across platforms \cite{SJ19}; the approach incorporated the Search using Random Forests (SuRF). 
SuRF is the earliest version of the parameter space search ytopt \cite{YTO} and only supports random forests. Currently, ytopt leverages Bayesian optimization \cite{BH19,SL12} to explore the parameter space search and uses different supervised learning methods within Bayesian optimization such as random forests, Gaussian process regression, extra trees, and gradient boosted regression trees. Some recent work has used machine learning and sophisticated statistical learning methods to reduce the overhead of autotuning \cite{RB16, MA17, TN18}. 

Kruse and Finkel \cite{KF19} implemented a newly proposed prototype of user-directed loop transformations using Clang \cite{clang} and Polly \cite{POLL} with additional loop transformation pragmas such as loop reversal, loop interchange, tiling, and array packing in DOE's Exascale Computing Project (ECP) SOLLVE  \cite{SOLL}. SOLLVE seeks to deliver a high-quality, robust implementation of OpenMP and project extensions in LLVM \cite{LLVM}, which is a collection of modular and reusable compiler and toolchain technologies. Research is needed to determine how to efficiently combine these loop transformation pragmas to optimize an application. Because of the large parameter space of these pragmas and related parameters, autotuning for automatic exploration of the parameter space is desirable. In our preliminary work with the Y-TUNE project \cite{WK20}, we worked on integrating the loop optimization pragmas within the ytopt package to autotune loop optimization pragmas for optimal performance. In this paper, we develop an autotuning framework to integrate the Clang/Polly loop optimization pragmas with the ytopt, and we apply the loop optimization pragmas to the PolyBench benchmarks \cite{YP16} to evaluate the autotuning framework.

PolyBench 4.2 \cite{YP16} is a benchmark suite of 30 numerical computations extracted from operations in various application domains (linear algebra computations, image processing, physics simulation, and data mining). In this work, we select six of the most complex benchmarks from the application domains of PolyBench benchmarks (syr2k, 3mm, heat-3d, lu, covariance, and Floyd-Warshall) and apply the newly developed LLVM Clang/Polly loop optimization pragmas to these benchmarks to improve their performance. 

We evaluate the performance on a machine with 3.1 GHz Quad-core Intel Core i7 and 16 GB of memory. The experimental results show that the autotuning outperforms the other compiling methods to provide the smallest execution time for the benchmarks syr2k, 3mm, heat-3d, lu, and covariance with two large datasets in 200 evaluations for effectively searching the parameter spaces with up to 170,368 different configurations. We compare four different supervised ML methods within Bayesian optimization and evaluate their effectiveness. We find that one exception for Polly is the Floyd-Warshall benchmark because Polly uses heuristics to optimize the benchmark to make it run much slower. To cope with this issue, we provide some compiler option solutions to improve the performance. Then we present loop autotuning without a user's knowledge using a simple mctree autotuning framework to further improve the performance of the Floyd-Warshall benchmark. We also extend the ytopt autotuning framework to tune a deep learning application.

This paper makes the following contributions:
 \begin{itemize}
\item We leverage ytopt autotuning framework to explore the parameter space search with the newly developed Clang loop optimization pragmas and compare four different supervised learning methods within Bayesian optimization and evaluate their effectiveness. 
\item We apply the loop optimization pragmas to the PolyBench benchmarks to optimize them.
\item We show that the autotuning framework outperforms other compiling methods to achieve the optimal implementation in 200 code evaluations for effectively searching the parameter spaces with up to 170,368 different configurations.
\item We present loop autotuning without a user's knowledge using a simple mctree autotuning framework to further improve the application performance.
\item We extend the ytopt autotuning framework to tune a deep learning application.
\end{itemize}

The remainder of this paper is organized as follows. Section 2 discusses SOLLVE Clang/Polly loop optimization pragmas and the parameter space search ytopt and then presents an autotuning framework based on them. Section 3 surveys the PolyBench benchmarks and selects six of the most complex benchmarks from the application domains. Section 4 applies the Clang loop optimization pragmas to these benchmarks to improve them and then use the autotuning framework to autotune the pragma parameters to achieve the optimal performance. We also compare four different supervised learning methods within Bayesian optimization and evaluate their effectiveness. Section 5 presents loop autotuning without a user's knowledge using a simple mctree autotuning framework to further improve the performance of the Floyd-Warshall benchmark. Section 6 extends the ytopt autotuning framework to tune a deep learning application. Section 7 discusses the related work, and Section 8 summarizes our conclusions and discusses future work. 


\section{The Autotuning Framework}

In this section, we discuss loop optimization pragmas implemented in LLVM Clang/Polly, the parameter search space, and the autotuning search method.

\subsection{Clang/Polly Loop Optimization Pragmas}

Compiler directives such as pragmas can help programmers to separate an algorithm's semantics from its optimization. Pragma directives for code transformations are useful for assisting program optimization and are already widely used in OpenMP. In~\cite{KF19}, a prototype of user-directed loop transformations using Clang and Polly~\cite{POLL} was implemented for the US DoE's ECP SOLLVE project~\cite{SOLL}.
Polly is LLVM's polyhedral loop optimizer which makes it easy to apply specific transformations as directed by pragmas.
We used the SOLLVE project's development branch for LLVM located at \url{https://github.com/SOLLVE/llvm-project/tree/pragma-clang-loop}.
While the SOLLVE team is working on integrating the changes into the official LLVM repository, only few of the changes have been upstreamed yet.
The additional loop transformation directives supported are loop reversal (inverting the iteration order of a loop), loop interchange (permuating the order of nested loops), tiling, unroll(-and-jam), array packing (temporarily copying the data of a loop's working set into a new buffer) and thread parallelization. More importantly, it supports composing multiple loop nest transformation in arbitrary order.
Vectorization is also supported by LLVM's dedicated loop vectorizer.
These pragmas are intended to make applying common loop optimization technique much easier and allow better separation of a code's semantics and its optimization. In this paper, we use some of these pragmas to optimize several PolyBench benchmarks and then propose the framework to autotune them.

\subsection{ytopt Autotuning System}

ytopt \cite{YTO} is an open source Python package that leverages Bayesian optimization \cite{BH19,SL12} to explore a user defined parameter space and uses different supervised ML methods within Bayesian optimization such as random forests, Gaussian process regression, extra trees, and gradient boosted regression trees. It is built on top of scikit-optimize \cite{SCIO} package and integrates ConfigSpace \cite{CFS} package to mange the algebraic constraints on the search spaces. See our initial work \cite{WK20} for the detailed installation and download information. ConfigSpace includes various modules to express different types of hyperparameters such as numerical, categorical, conditional and ordinal, and constraints such as conditions and forbidden clauses. ConfigSpace is often used in automated machine learning tools \cite{AML}.

\begin{figure}
\centering
\begin{minipage}{.5\textwidth}
  \centering
 \includegraphics[width=.55\textwidth]{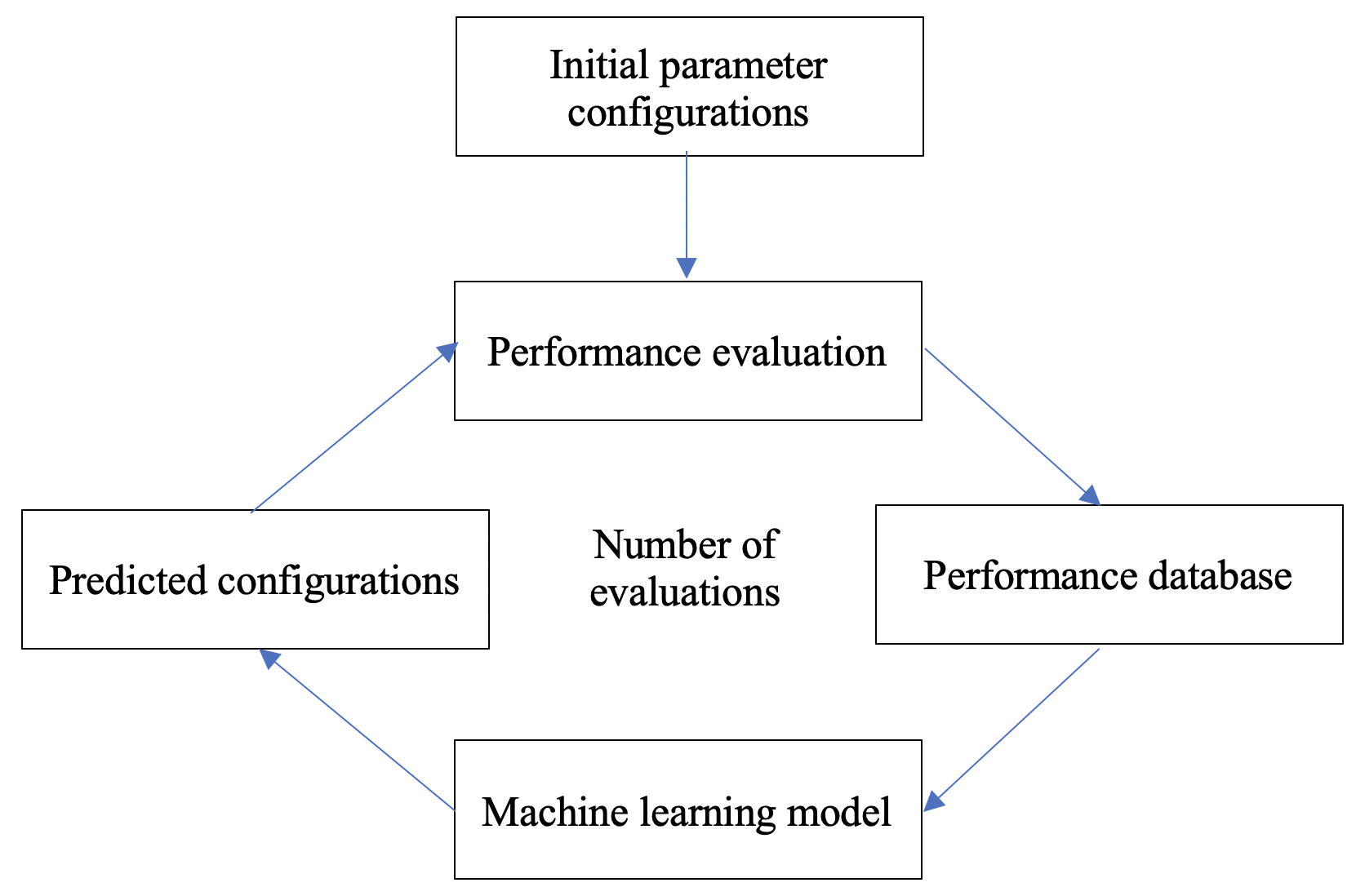}
 \caption{Machine-learning-based autotuner ytopt}
\label{fig1}
\end{minipage}%
\begin{minipage}{.5\textwidth}
  \centering
 \includegraphics[width=1\textwidth]{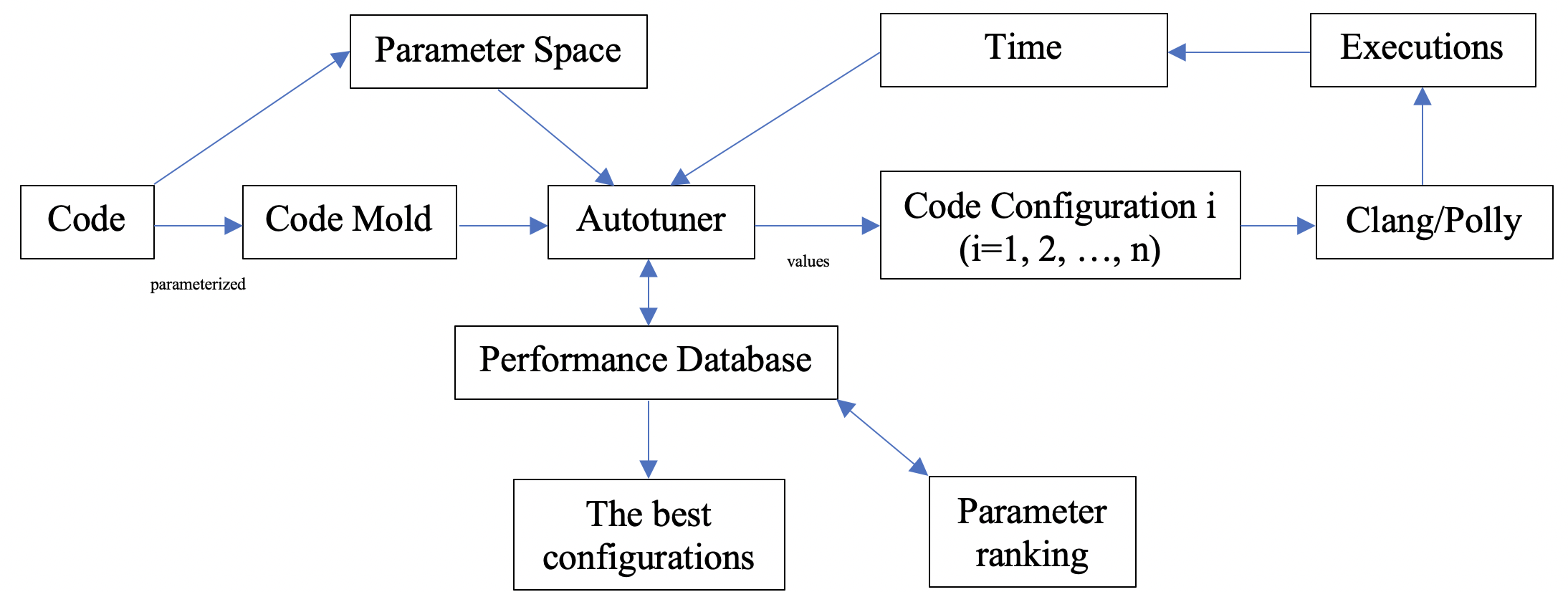}
 \caption{ytopt autotuning framework for Clang loop pragmas}
\label{fig2}
\end{minipage}
\end{figure}

A high level overview of ytopt is shown in Figure \ref{fig1}. It takes the user-defined parameter space definition (bounds and constraints) and the parameter configuration evaluation interface as input. The initialization phase consists of sampling  a  small  number  of  input  parameter configurations through random sampling or Latin hypercube sampling and recording the performance (runtime) to a performance database.
The configurations and their performance values are used to fit a surrogate model using a supervised learning method. A key requirement of the surrogate model is that it should provide a point estimate and predictive variance associated with the point estimate. The iterative phase of search consists in sampling an input parameter configuration from the parameter space described in ConfigSpace for evaluation by progressively leveraging and refining the surrogate model. It starts by constructing a batch of valid configurations, which are randomly sampled from the ConfigSpace object. The configurations that are already evaluated are removed from the batch. A configuration for evaluation is selected from the generated batch as follows. For each configuration $x_M^i$ in the batch, the trained model $M$ is used to predict a point estimate (mean value) $\mu(x_M^i)$ and standard deviation $\sigma(x_M^i)$. These configurations are ranked by using the lower confidence bound (LCB) acquisition function:
\begin{equation}
    a_{LCB}(x_M^i) = \mu(x_M^i) - \kappa\sigma(x_M^i),
    \label{eqn:lcb}
\end{equation}
where $\kappa \geq 0$ is a user-defined parameter that controls the tradeoff between exploration and exploitation. When $\kappa=0$, the search will be characterized by pure exploitation, where the configuration with the lowest mean value is always selected. When $\kappa$ is set to a large value, the search will perform pure exploration, where a configuration with large predictive variance is always selected. Evaluation of such configurations results in improvement of the model $M$. 
In brief, the search uses Bayesian optimization in which uncertainty quantification of the surrogate model is leveraged to balance exploration of the search space and identification of more-promising regions using LCB acquisition function.

The main difference between the current ytopt and the previous version SuRF is the support for algebraic constraints and a search method that works directly over the valid configurations. This is achieved by integrating ConfigSpace library within ytopt and modification of random sampler to use the ConfigSpace random sampler that generates only valid configurations. 
The ytopt supports four supervised ML methods: random forests, Gaussian process regression, extra trees, and gradient boosted regression trees. Three of the four ML methods---random forest, Extra Trees, and gradient boosted regression trees---follow the autotuning process based on the performance database shown in Figure \ref{fig1}. Gaussian process, however, still uses random sampling to generate the parameter configurations for performance evaluation.

\subsection{Proposed ytopt Autotuning Framework}

Based on the Clang loop optimization pragmas and the parameter space search ytopt, we present the ytopt autotuning framework in the following steps shown in Figure \ref{fig2}:

 \begin{itemize}
\item [1)] Analyze an application code to identify the important parameters to focus on.
\item	[2)] Replace these parameters with marker symbols such as \#P0, \#P1, \#P2, ..., \#Pm in  top-down order to generate another code with these symbols as a code mold. 
\item	[3)] Define the value ranges of these symbols for the parameter space as an input of the ytopt autotuner (problem.py).
\item	[4)] Use the ytopt autotuner to search the parameter space to select the values in the allowed ranges (using random forest as default), and replace these symbols in the mold code with them to generate a new code using the function plopper (plopper.py).
\item	[5)] Compile the new code and execute it to measure the execution time (using a Perl script exe.pl).
\item	[6)] Record the execution time and the elapsed time with the parameters' values to the performance database (two output files: results.csv and results.json).
\item	 [7)] Repeat steps 4, 5, and 6 until  reaching the maximum number of code evaluations n (default: 100).
\item	[8)] Process the database to find the smallest execution time and output the optimal configuration for the execution time (findMin.py)
\item [9)] Identify the most important features which impact the performance for the search improvement.
 \end{itemize}

This autotuning framework requires the following components: configspace, scikit-optimize, autotune, ytopt, and LLVM clang/polly. The framework provides the following main options: 
 \begin{itemize}
\item[]  --max-evals: to set the maximum number of evaluations n
\item[]  --learner: to set the ML method using random forests (RF),  Extra Trees (ET), gradient Boosted regression trees (GBRT), or Gaussian processes (GP). The default is RF.
\item[]  --evaluator:  to set the evaluator using balsam, subprocess, or ray method. The default is ray method.
\item[]  --eval-timeout-minutes: to set the timeout minutes 
\end{itemize}

In this work, we apply the Clang loop optimization pragmas to the PolyBench benchmarks \cite{YP16} to evaluate the ytopt autotuning framework using the four different ML methods.

\section{PolyBench Benchmarks}

PolyBench 4.2 \cite{YP16} is a benchmark suite of 30 numerical computations extracted from operations in various application domains (19 linear algebra computations, 3 image-processing applications, 6 physics simulations, and 2 data-mining applications). The details about the benchmarks are as follows:
 \begin{itemize}
\item	[1)] Linear Algebra: 

BLAS (7):  gemm,	gemver,	 gesummv,	symm,	\textbf{syr2k},	syrk,	trmm

Kernels (6): 2mm,	\textbf{3mm},	atax,	bicg,	doitgen, 	mvt

Solvers (6): Cholesky,  durbin,  gramschmidt,	 \textbf{lu},	ludcmp,		trisolv

\item	[2)] Medley (image processing) (3): deriche,	\textbf{floyd-warshall},	  nussinov

\item	[3)] Physics simulation (stencils) (6): adi,  fdtd-2d,  \textbf{heat-3d},  jacobi-1d,  jacobi-2d,  seidel-2d

\item	[4)] Data mining (2): correlation,  \textbf{covariance}
\end{itemize}

In this work, we choose the most complex benchmark with the most levels of nested loops (in bold) from each group to illustrate how the autotuning framework performs, and we compare their performance.

syr2k is a symmetric rank 2k update from BLAS and entails the matrix multiplication C = A*alpha*B+ B*alpha*A+belta*C, where A is an NxM matrix, B is an MxN matrix, and C is an NxN symmetric matrix. We use the following large datasets: LARGE\_DATASET (M 1000, N 1200) and EXTRALARGE\_DATASET (M 2000, N 2600) for our case study.

3mm is one of the linear algebra kernels that consists of three matrix multiplications and entails G=(A*B)*(C*D), where A is a PxQ matrix; B is a QxR matrix; C is an RxS matrix; and D is an SxT matrix. We use the following large datasets: LARGE\_DATASET (P 800, Q 900, R 1000, S 1100, T 1200) and EXTRALARGE\_DATASET (P 1600, Q 1800, R 2000, S 2200, T 2400).

lu is LU decomposition without pivoting in linear algebra solvers and entails A = L*U, where L is an NxN lower triangular matrix and U is an NxN upper triangular matrix. We use the following large datasets: LARGE\_DATASET (N 2000) and EXTRALARGE\_DATASET (N 4000).

heat-3d entails a heat equation over 3D space in Stencil. Stencil computations iteratively update a grid of data using the same pattern of computation. We use the following large datasets: LARGE\_DATASET (TSTEPS 500, N 120) and EXTRALARGE\_DATASET (TSTEPS 1000, N 200).

covariance entails computing the covariance, a measure from statistics that shows how linearly related two variables are. It takes the data (NxM matrix that represents N data points, each with M attributes) as input and gives the cov (symmetric MxM matrix where the i,jth element is the covariance between i and j) as the output. We use the following large datasets: LARGE\_DATASET (M 1200, N 1400) and EXTRALARGE\_DATASET (M 2600, N 3000).

Floyd-Warshall entails computing the shortest paths between each pair of nodes in a graph in Medley. The input is an NxN matrix, where the i,jth entry represents the cost of taking an edge from i to j. The output is an NxN matrix, where the i,jth entry represents the shortest path length from i to j.  We use the following datasets: MEDIUM\_DATASET (N 500) and LARGE\_DATASET (N 2800).

\section{Autotuning the Benchmarks with Clang Loop Optimization Pragmas}

We apply the Clang loop optimization pragmas \cite{KF19} to the chosen PolyBench benchmarks to optimize them. We define the parameters for these pragmas and then apply the ytopt autotuning framework to these benchmarks. Based on the performance database, we find the smallest execution time and output the best configurations. We also use four different ML methods to investigate how they impact finding the optimal configuration from the input parameter search space. We evaluate the performance on a machine with 3.1 GHz Quad-core Intel Core i7 and 16 GB of memory and 1 TB SSD; gcc 7.2 and clang 10.0 are installed on the machine.

\subsection{Case Study: syr2k with multiple loop transformations (loop tiling, interchange, and array packing)}

We apply multiple loop transformations such as tiling, interchange, and array packing pragmas to the benchmark syr2k for this case study. We assume that the 3D loop tiling (\#pragma clang loop(i,j,k) tile sizes( )) is already applied to syr2k. We define the following parameters:

{\scriptsize
\begin{verbatim}
#P0
#P1
#P2
#pragma clang loop(i,j,k) tile sizes(#P3,#P4,#P5) floor_ids(i1,j1,k1) tile_ids(i2,j2,k2)
#pragma clang loop id(i)
  for (i = 0; i < _PB_N; i++) {
    #pragma clang loop id(j)
    for (j = 0; j < _PB_M; j++) {
     #pragma clang loop id(k)
        for (k = 0; k <= i; k++)
        {
          C[i][k] += A[k][j]*alpha*B[i][j] + B[k][j]*alpha*A[i][j];
        }
    }
  }
\end{verbatim}
 }  
 
Based on these parameters, we create a code mold. For this case, we have to make sure that there is no dependence among P0, P1, P2, P3, P4, and P5. We have the following parameter space input\_space using ConfigSpace:

{\scriptsize
\begin{verbatim}
cs  CS.ConfigurationSpace(seed=1234)
P0=CSH.CategoricalHyperparameter(name='P0', choices=["#pragma clang loop(j2) pack array(A) 
        allocate(malloc)", " "], default_value=' ')
P1=CSH.CategoricalHyperparameter(name='P1', choices=["#pragma clang loop(i1) pack array(B) 
        allocate(malloc)", " "], default_value=' ')
P2=CSH.CategoricalHyperparameter(name='P2', choices=["#pragma clang loop(i1,j1,k1,i2,j2) 
        interchange permutation(j1,k1,i1,j2,i2)", " "], default_value=' ')
P3=CSH.OrdinalHyperparameter(name='P3', sequence=['4','8','16','20','32','50','64','80','96','100','128'], 
        default_value='96')
P4=CSH.OrdinalHyperparameter(name='P4', sequence=['4','8','16','20','32','50','64','80','100','128','2048'], 
        default_value='2048')
P5=CSH.OrdinalHyperparameter(name='P5', sequence=['4','8','16','20','32','50','64','80','100','128','256'], 
        default_value='256')
cs.add_hyperparameters([P0, P1, P2, p3, P4, P5])
cond1 = CS.InCondition(P1, P0, ['#pragma clang loop(j2) pack array(A) allocate(malloc)'])
cs.add_condition(cond1)
input_space = cs 
\end{verbatim}
 }
 
where  the parameters P0, P1, and P2 have the choices of the pragmas or nothing. For instance, P0 has the choice of the array packing for A (\#pragma clang loop(j2) pack array(A) allocate(malloc)) or nothing. P2 permutes the order of nested loops and has no dependence with P0 and P1. For P0 and P1, we add the conditions (CS.InCondition)  which limit the parameters associated with a variant once a parameter choice makes others no longer relevant so that Packing arrays A and B occurs at the same time.
The parameters P3, P4, and P5 represent the tile size for each loop. Based on the tile size settings in \cite{LI16, BU20} related to the cache sizes, we set the default size to 96 for P3, 2048 for P4, and 256 for P5. For simplicity, we set 11 tile size choices for each of these parameters. Therefore, the parameter space consists of 2x2x2x11x11x11= 10,648 different configurations. Then we use our autotuning framework to find out which configuration results in the smallest execution time.

We use four ML methods---RF, ET, GBRT, and GP---to investigate which one generates the smallest runtime for which configuration in 200 evaluations for syr2k with the large dataset. We then use the best method to autotune the benchmark with the extra large dataset. 

\begin{figure}
\centering
\begin{minipage}{.5\textwidth}
  \centering
 \includegraphics[width=.6\textwidth]{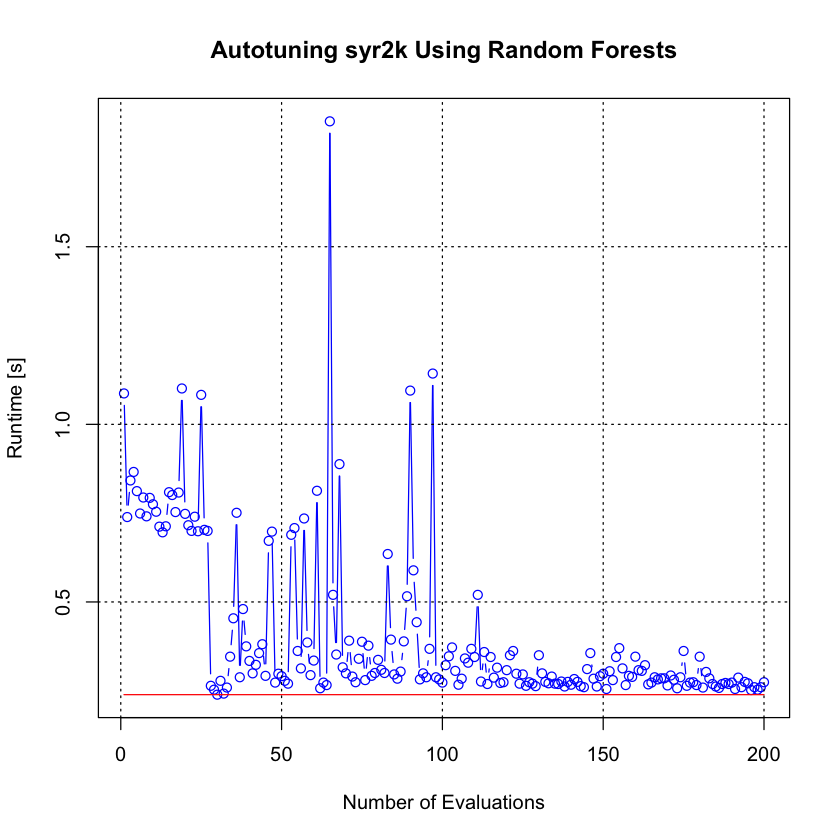}
 \caption{Autotuning syr2k using RF in 200 evaluations}
\label{fig3}
\end{minipage}%
\begin{minipage}{.5\textwidth}
  \centering
  \includegraphics[width=.6\textwidth]{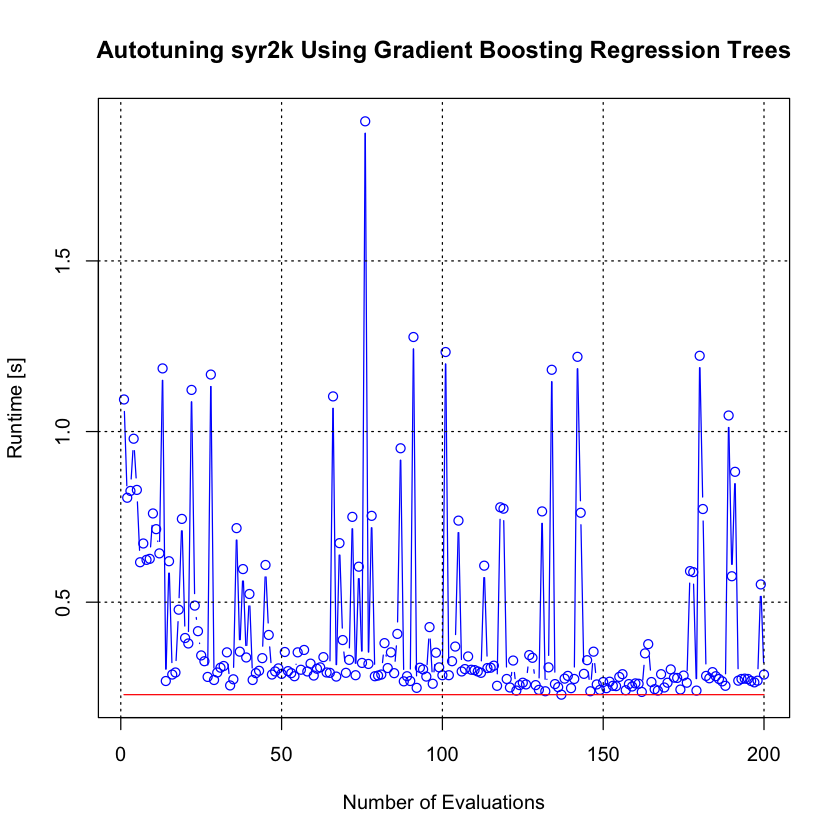}
 \caption{Autotuning syr2k using GBRT in 200 evaluations}
\label{fig4}
\end{minipage}
\end{figure}

In Figure \ref{fig3}, RF results in the smallest runtime of 0.239s for the configuration ('\#pragma clang loop(j2) pack array(A) allocate(malloc)','\#pragma clang loop(i1) pack array(B) allocate(malloc)','\#pragma clang loop(i1,j1,k1,i2,j2) interchange permutation(j1,k1,i1,j2,i2)',128,128,100 ) at Evaluation 30 of 200 evaluations. The blue line is for all evaluations; the red line is the best execution time. RF results in the execution time close to the smallest one with increasing the number of evaluations.

In Figure \ref{fig4}, GBRT results in the smallest runtime of 0.229s for the configuration ('\#pragma clang loop(j2) pack array(A) allocate(malloc)','\#pragma clang loop(i1) pack array(B) allocate(malloc)','\#pragma clang loop(i1,j1,k1,i2,j2) interchange permutation(j1,k1,i1,j2,i2)',50,128,256) at Evaluation 137 of 200 evaluations. 

\begin{figure}
\centering
\begin{minipage}{.5\textwidth}
  \centering
  \includegraphics[width=.6\textwidth]{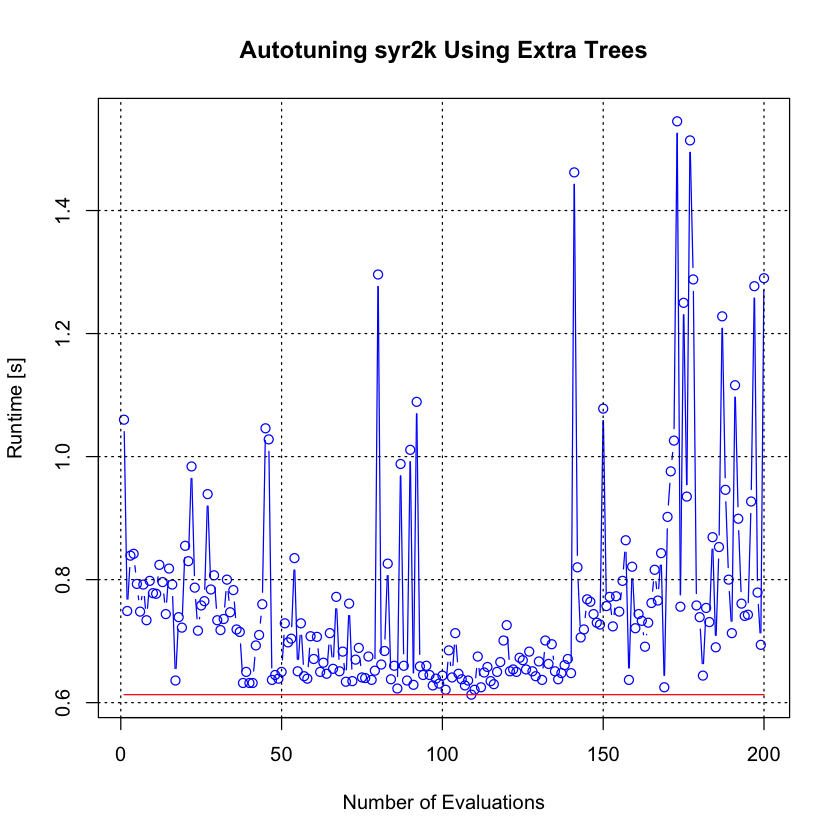}
 \caption{Autotuning syr2k using ET in 200 evaluations}
\label{fig5}
\end{minipage}%
\begin{minipage}{.5\textwidth}
  \centering
 \includegraphics[width=.6\textwidth]{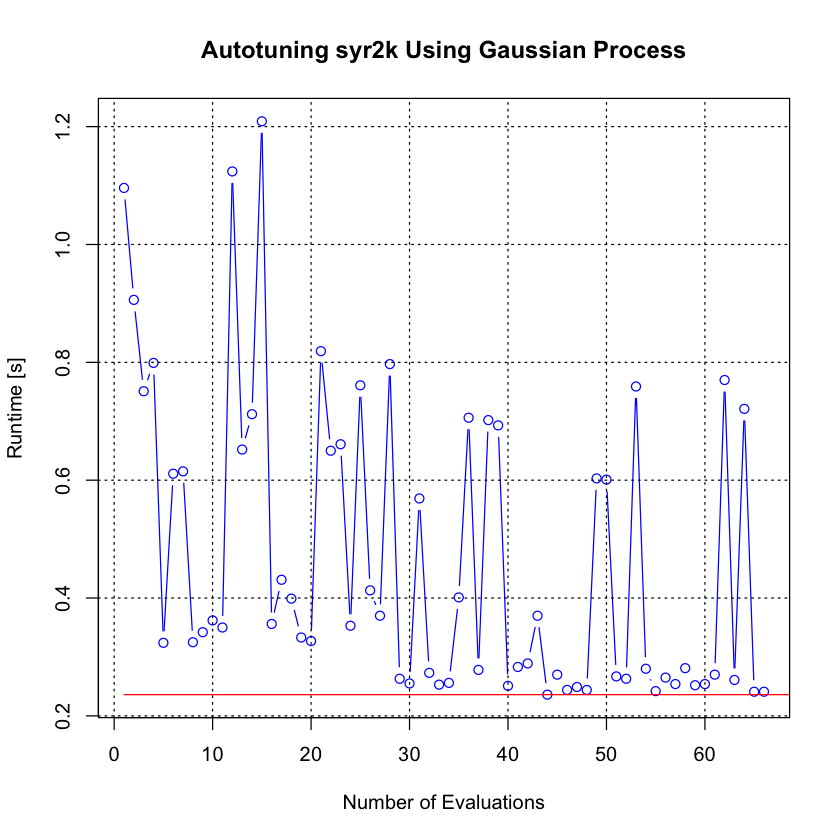}
 \caption{Autotuning syr2k using GP in 200 evaluations}
\label{fig6}
\end{minipage}
\end{figure}

In Figure \ref{fig5}, ET results in the smallest runtime of 0.613s for the configuration (' ', ' ', ' ' ,100,8,8 ) at Evaluation 109 of 200 evaluations.  ET does not show any pragmas in the best configuration with the only tile size (100, 8, 8).

In Figure \ref{fig6}, GP results in the smallest runtime of 0.236s for the configuration ('\#pragma clang loop(j2) pack array(A) allocate(malloc)','\#pragma clang loop(i1) pack array(B) allocate(malloc)','\#pragma clang loop(i1,j1,k1,i2,j2) interchange permutation(j1,k1,i1,j2,i2)',80,100,256 ) at Evaluation 44 and finishes only 66 evaluations. As  discussed in Section 2.2, GP does not use the performance database to assist the parameter space search as designed, and thus it uses only 66 of the 200 evaluations. The other 134 evaluations are skipped because of the replicated evaluations. The other methods RF, GBRT, and ET use the performance database to assist the parameter space search, and they finish all 200 evaluations with different configurations.

\begin{table}
\center
\caption{Performance (in seconds) comparison of syr2k using different compilers and autotuning}
\begin{tabular}{c}
  \includegraphics[width=.47\textwidth]{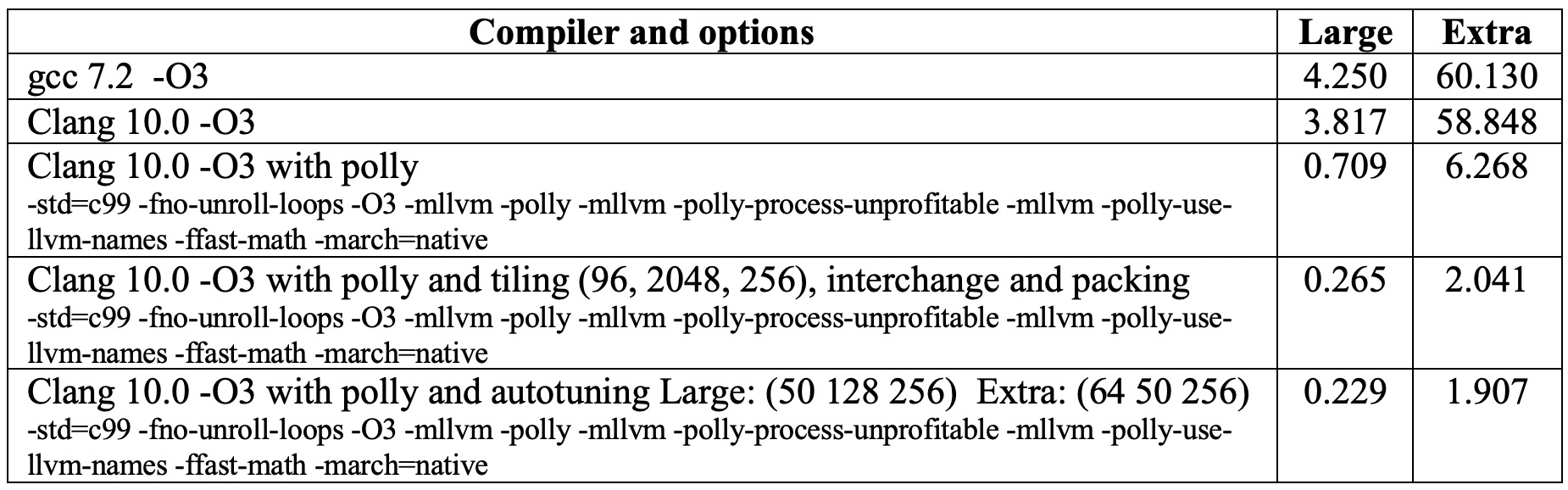}
  \end{tabular}
\label{tab1}       
\end{table} 

Table \ref{tab1} shows the performance comparison of syr2k using different compilers and autotuning. For the first three rows, these compilers and options are applied to the original code syr2k (without any loop pragma) to measure the smallest execution time in 10  runs. The fourth rows are for the syr2k with the loop tiling and default tile size (96, 2048, 256), interchange, and array packing. The last row is the results using autotuning. The best configuration in 200 evaluations has the tile size (50 128 256) for the large dataset and the tile size (64 50 256) for the extra large dataset using GBRT. We observe that autotuning outperforms the other compiling methods to provide the smallest execution time for both datasets. 

\subsection{Case Study: 3mm with multiple loop transformations} 
\begin{figure}
\centering
\begin{minipage}{.5\textwidth}
  \centering
 \includegraphics[width=.6\textwidth]{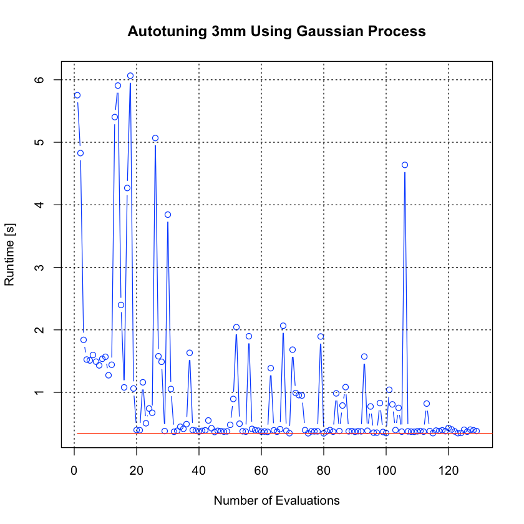}
 \caption{Autotuning 3mm using GP in 200 evaluations}
\label{fig7}
\end{minipage}%
\begin{minipage}{.5\textwidth}
  \centering
 \includegraphics[width=.6\textwidth]{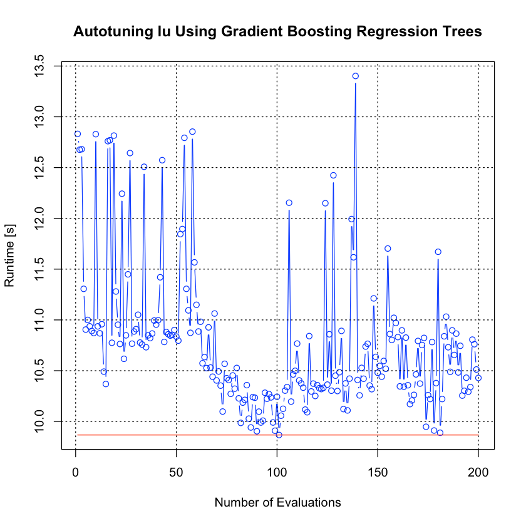}
 \caption{Autotuning lu using GBRT in 200 evaluations}
\label{fig8}
\end{minipage}
\end{figure}

We apply multiple loop transformations such as tiling, interchange, and array packing pragmas to the benchmark 3mm for this case study. We define 10 pragmas parameters to autotune the benchmark with the parameter space of 170,368 different configurations and use four ML methods---RF, ET, GBRT, and GP---to investigate which one generates the smallest runtime for which configuration in 200 evaluations for 3mm with the large dataset. We find that GP results in the smallest runtime of 0.345 s for the configuration (' ', ' ', ' ', 80, 100, 4, ' ', ' ', ' ' ,' ') at Evaluation 80 and finishes 129 evaluations as shown in Figure \ref{fig7}. Then we use GP to autotune the benchmark for the extra large dataset.

\begin{table}
\center
\caption{Performance (in seconds) comparison of 3mm using different compilers and autotuning}
\begin{tabular}{c}
  \includegraphics[width=.47\textwidth]{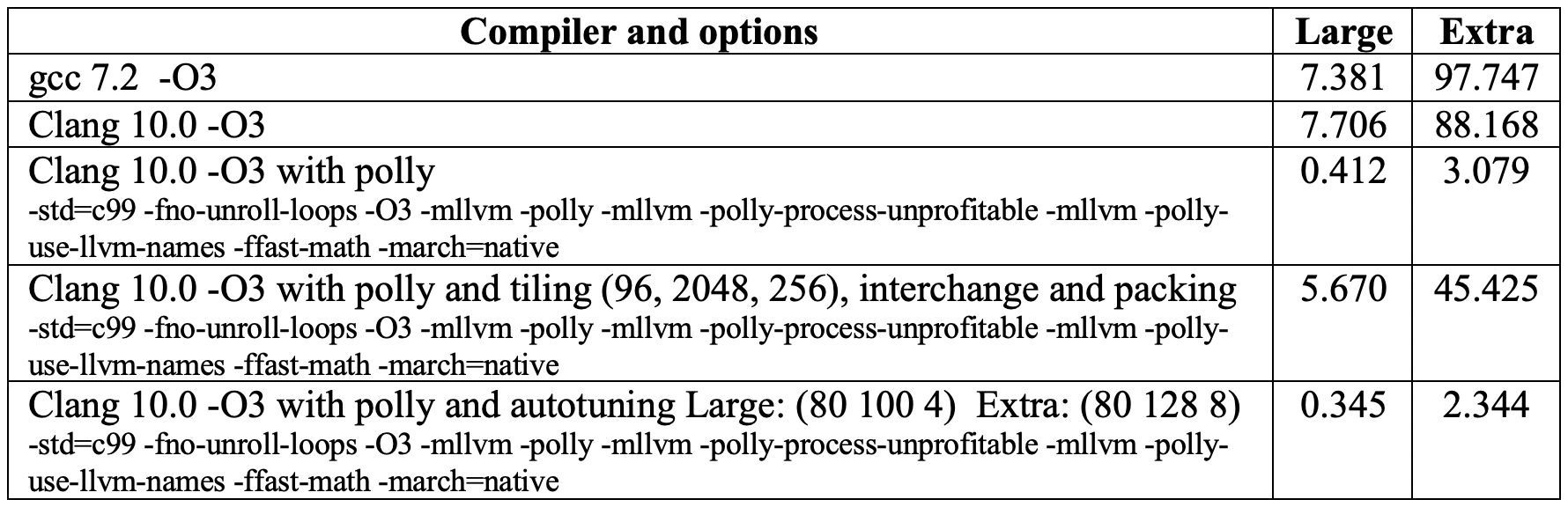}
  \end{tabular}
\label{tab2}       
\end{table} 

Table \ref{tab2} shows the performance comparison of 3mm using different compilers and autotuning. For the first three rows, these compilers and options are applied to the original code 3mm (without any loop pragma) to measure the smallest execution time in 10 same runs. The fourth rows are for the 3mm with the loop tiling with default tile size (96, 2048, 256), interchange, and array packing. The last row is the results using autotuning. The best configuration in 200 evaluations has the tile size (80 100 4)  for the large dataset and the tile size (80 128 8) for the extra large dataset using GP. We observe that autotuning outperforms the other methods to provide the smallest execution time for both datasets. 

\subsection{Case Study: lu with multiple loop transformations}

We apply multiple loop transformations such as tiling, interchange, and array packing pragmas to the benchmark lu for this case study. We define the five pragma parameters to autotune the benchmark. We use four ML methods---RF, ET, GBRT, and GP---to check which one generates the smallest runtime for which configuration in 200 evaluations for lu with the large dataset. We find that GBRT has the smallest runtime of 9.867 s for the configuration ('\#pragma clang loop(i1) pack array(A) allocate(malloc)', ' ', 50, 2048, 8) at Evaluation 101 of 200 evaluations, as shown in Figure \ref{fig8}. Then we use GBRT to autotune the benchmark with the extra large dataset.

\begin{table}
\center
\caption{Performance (in seconds) comparison of lu using different compilers and autotuning}
\begin{tabular}{c}
  \includegraphics[width=.47\textwidth]{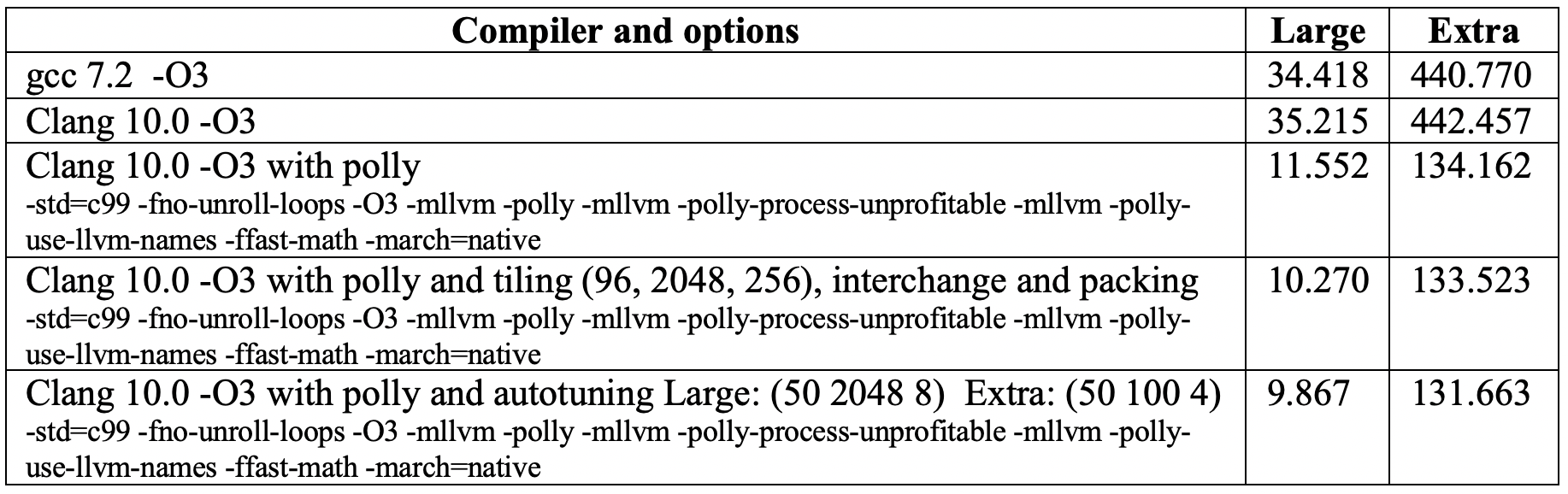}
  \end{tabular}
\label{tab3}       
\end{table} 

Table \ref{tab3} shows the performance comparison of lu using different compilers and autotuning. We find that the best configuration in 200 evaluations has the tile size (50 2048 8) for the large dataset and the tile size (50 100 4) for the extra large dataset using GBRT. We observe that autotuning outperforms the other methods to provide the smallest execution time for both datasets. 

\subsection{Case Study: heat-3d with multiple loop transformations} 

\begin{figure}
\centering
\begin{minipage}{.5\textwidth}
  \centering
 \includegraphics[width=.6\textwidth]{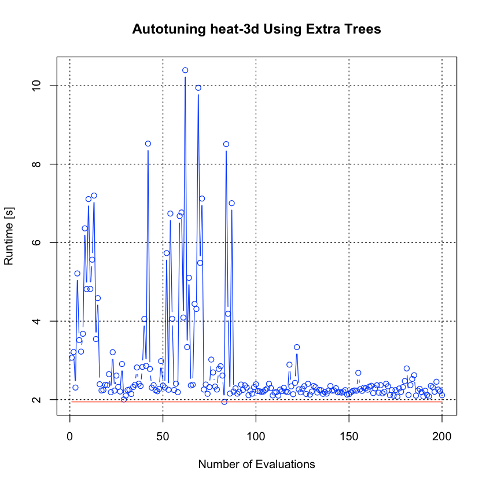}
 \caption{Autotuning heat-3d using ET in 200 evaluations}
\label{fig9}
\end{minipage}%
\begin{minipage}{.5\textwidth}
  \centering
 \includegraphics[width=.6\textwidth]{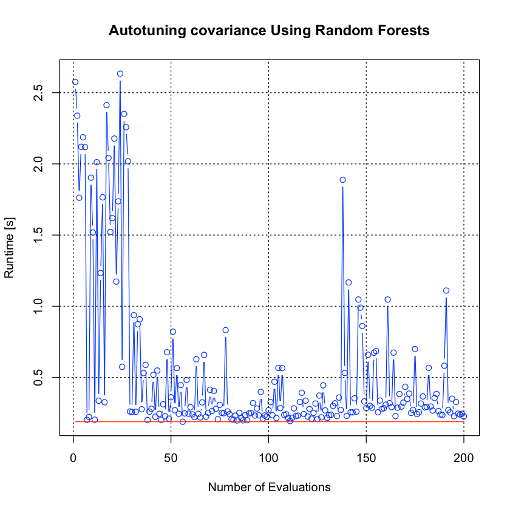}
 \caption{Autotuning covariance using RF in 200 evaluations}
\label{fig10}
\end{minipage}
\end{figure}

We apply multiple loop transformations such as tiling, interchange, and array packing pragmas to the benchmark heat-3d for this case study. We define the six pragma parameters to autotune the benchmark. We use four ML methods---RF, ET, GBRT, and GP---to check which one generates the smallest runtime for which configuration in 200 evaluations for heat-3d with large dataset. We find that ET has the smallest runtime of 1.942 s for the configuration (' ', ' ' ,' ', 100, 64, 128) at Evaluation 83 of 200 evaluations, as shown in Figure \ref{fig9}. We then use ET to autotune the benchmark with the extra large dataset.

\begin{table}
\center
\caption{Performance (in seconds) comparison of heat-3d using different compilers and autotuning}
\begin{tabular}{c}
  \includegraphics[width=.47\textwidth]{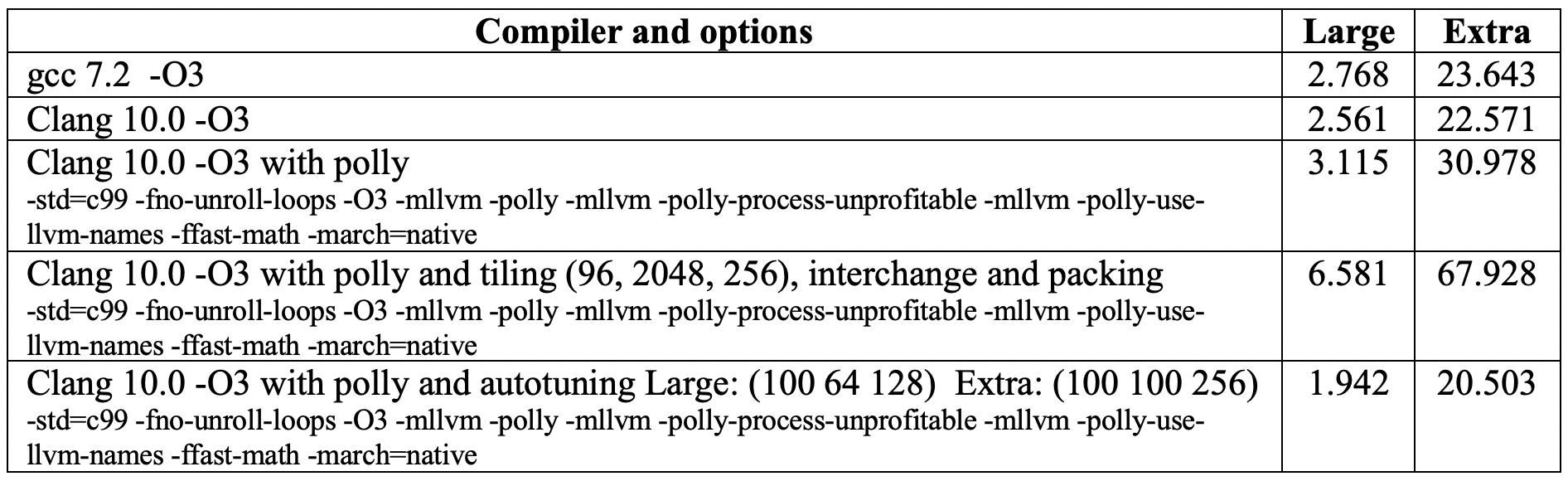}
  \end{tabular}
\label{tab4}       
\end{table} 

Table \ref{tab4} shows the performance comparison of heat-3d using different compilers and autotuning. We find that the best configuration in 200 evaluations has the tile size (100 64 128)  for the large dataset and the tile size (100 100 256) for the extra large dataset using ET. We observe that autotuning outperforms the other methods to provide the smallest execution time for both datasets. 

\subsection{Case Study: covariance with multiple loop transformations} 

We apply multiple loop transformations such as tiling, interchange, and array packing pragmas to the benchmark covariance for this case study. We define the five pragma parameters to autotune the benchmark. We use four ML methods---RF, ET, GBRT, and GP---to check which one generates the smallest runtime for which configuration in 200 evaluations for covariance with large dataset. We  find that RF has the smallest runtime of 0.188s for the configuration (' ' ,' ', 96, 100, 8 ) at Evaluation 56 of 200 evaluations, as shown  in Figure \ref{fig10}. We then use RF to autotune the benchmark with the extra large dataset.

\begin{table}
\center
\caption{Performance (seconds) comparison of covariance using different compilers and autotuning}
\begin{tabular}{c}
  \includegraphics[width=.47\textwidth]{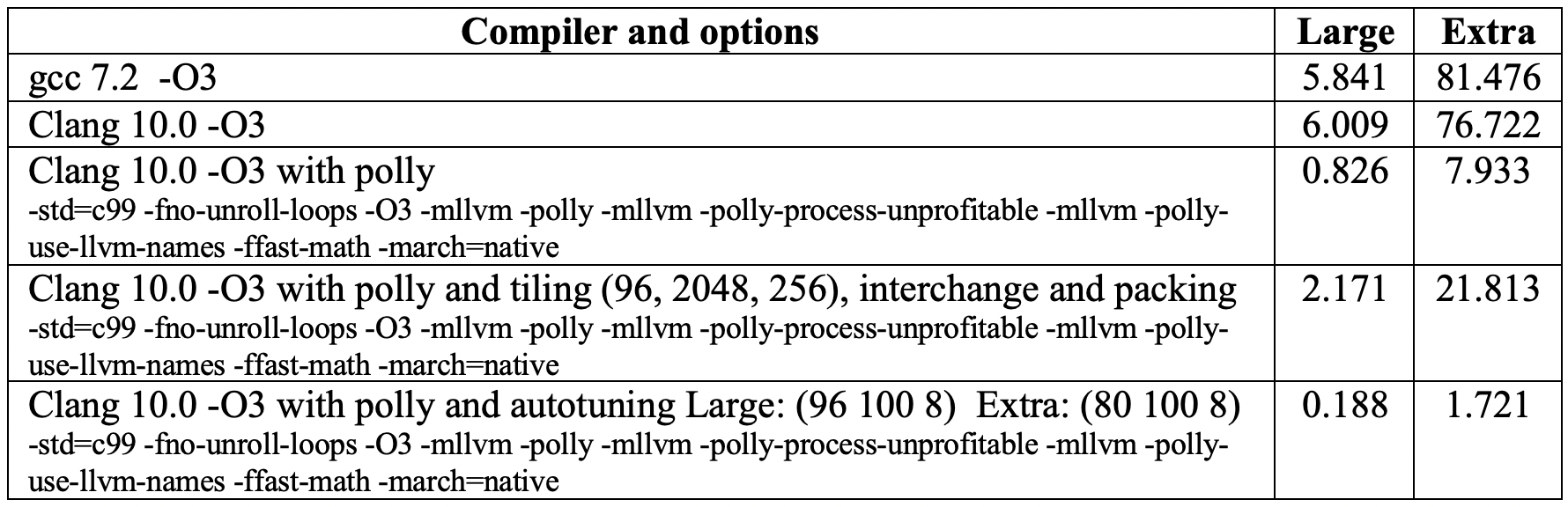}
  \end{tabular}
\label{tab5}       
\end{table} 

Table \ref{tab5} shows the performance comparison of covariance using different compilers and autotuning. We find that the best configuration in 200 evaluations has the tile size (96 100 8)  for the large dataset and the tile size (80 100 8) for the extra large dataset using RF. We observe that autotuning outperforms the other methods to provide the smallest execution time for both datasets. 

\subsection{Case Study: Floyd-Warshall with multiple loop transformations}

We apply multiple loop transformations such as tiling, interchange, and array packing pragmas to the benchmark Floyd-Warshall for this case study. We find that when we compiled the code, the following warning occurred: "floyd-warshall.c:89:5: warning: loop(s) not tiled: transformation would violate dependencies [-Wpass-failed=polly-opt-isl]." That is, the pragmas were ineffective, and Polly applied its default transformation. When we ran the benchmark with the large dataset, it takes more than 135 s shown in Table \ref{tab6}. We note that Clang with Polly causes the benchmark to run very slowly (almost 9 times slower). We investigate what happened as follows.

\begin{table}
\center
\caption{Performance comparison of Floyd-Warshall using different compilers and options}
\begin{tabular}{c}
  \includegraphics[width=.47\textwidth]{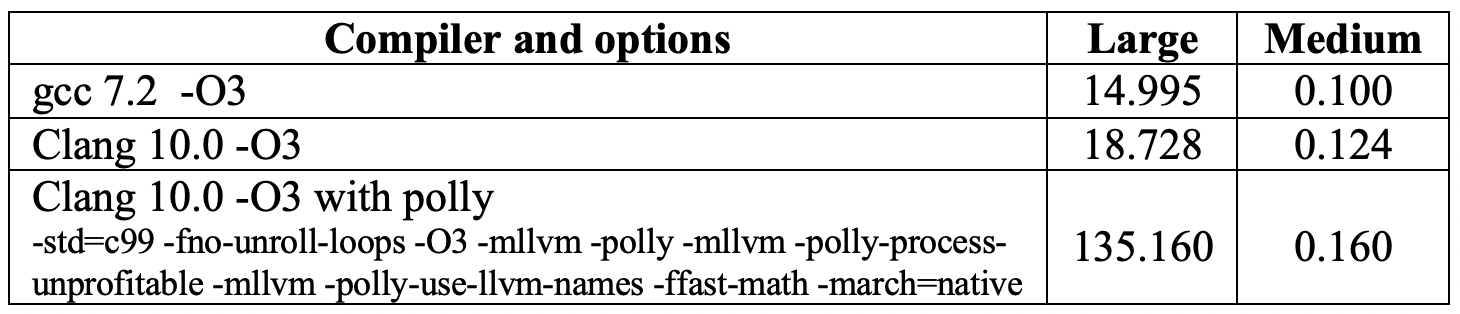}
  \end{tabular}
\label{tab6}       
\end{table}

In the PolyBench Floyd-Warshall code, the main kernel looks like the following.
{\scriptsize
\begin{verbatim}
for (k = 0; k < _PB_N; k++)
  for(i = 0; i < _PB_N; i++)
    for (j = 0; j < _PB_N; j++)
      path[i][j] = (path[i][j] < path[i][k] + path[k][j]) ? path[i][j]: (path[i][k] + path[k][j]);
\end{verbatim}
 }
Most accesses need to have the innermost subscript dependent on the fastest iterating induction variable j, meaning consecutive memory accesses in the innermost loop (i.e., spatial locality) enable effective use of cache lines and prefetching by the CPU.

Using Polly's default loop optimization heuristic implemented by ISL (Integer Set Library), the following schedule is applied.

{\scriptsize
\begin{verbatim}
for (c0 = 0; c0 <= 2799; ++c0)
  for (c1 = 0; c1 <= 5598; ++c1)
    for (c2 = max(0, c1 - 2799); c2 <= min(2799, c1); ++c2)
      Stmt_for_body6_i(c0, c2, c1 - c2);
\end{verbatim}
 }

This corresponds to something like the following loop.
{\scriptsize
\begin{verbatim}
for (j = 0; j < _PB_N; j++)
  for (k = 0; k < _PB_N; k++)
    for(i = _PB_N-1; i >= 0; i--)
      path[i][j] = (path[i][j] < path[i][k] + path[k][j]) ? path[i][j]: (path[i][k] + path[k][j]);
\end{verbatim}
 }
 
In this variant, the fastest-running index i is not the innermost subscript of any of the array accesses. In other words, all the accesses are strided in memory and will access different cache lines from those of the previous iterations. This is bad for performance.
The default loop nest optimization strategy only considers temporal reuse, but not spatial reuse. Therefore, it does not prioritize keeping the j-loop as the innermost loop with the fastest-running index, making the execution slower than the original loop nest.

\begin{table}
\center
\caption{Performance improvement of Floyd-Warshall using Clang/Polly with additional options and autotuning}
\begin{tabular}{c}
  \includegraphics[width=.47\textwidth]{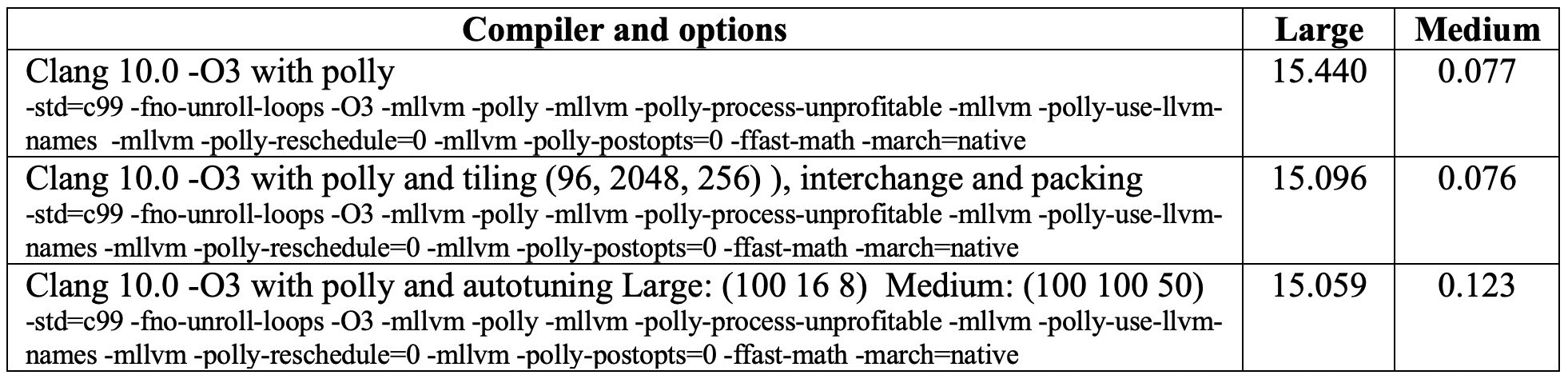}
  \end{tabular}
\label{tab7}       
\end{table} 

\begin{figure}
\center
 \includegraphics[width=.3\textwidth]{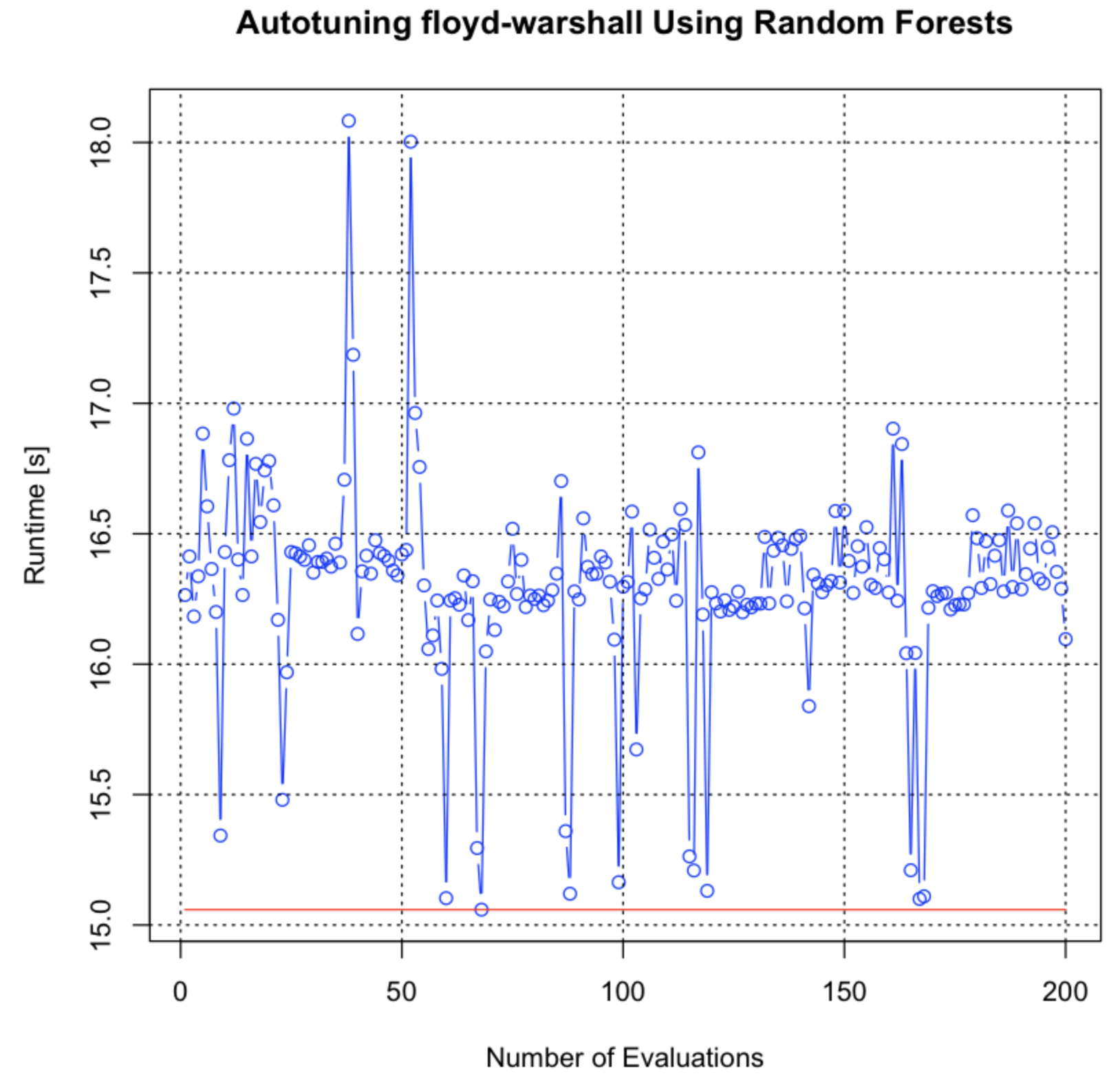}
 \caption{Autotuning Floyd-Warshall using RF in 200 evaluations}
\label{fig12}
\end{figure}

To cope with this situation, we updated \url{https://github.com/SOLLVE/llvm-project/tree/pragma-clang-loop} to include new flags: the flags -mllvm -polly-reschedule=0 -mllvm -polly-postopts=0. These flags make Polly do nothing if no pragma is applied, instead of using the default optimizer. Additionally, and the flag -mllvm -polly-pragma-ignore-depcheck makes Polly apply a transformation even if it cannot confirm that it is semantically correct. 
In this case it is necessary because the ternary operation applies a max-reduction, which is commutative but the computer cannot detect.
Hence, adding the flag forces Polly to apply tiling even though the compiler cannot ensure its semantic legality.
We find that RF has the smallest runtime of 15.059 s for the configuration (' ' ,' ', 100, 16, 8 ) at Evaluation 68 of 200 evaluations for the large dataset in Figure \ref{fig12}. We then use RF to autotune the benchmark with the medium dataset. Compared to Table \ref{tab6}, we observe that the significant performance improvement occurs from around 135s to around 15s in Table \ref{tab7} with Polly by adding these flags -mllvm -polly-reschedule=0 -mllvm -polly-postopts=0. Overall, this benchmark illustrates that a heuristic-based optimization can also regress a program's performance due to being unable to model the entire architecture complexity and unavailability of dynamic information such as the actual execution time used by the autotuning approach.

\section{Loop Autotuning without a User's Knowledge}

\begin{figure}
\center
 \includegraphics[width=.6\textwidth]{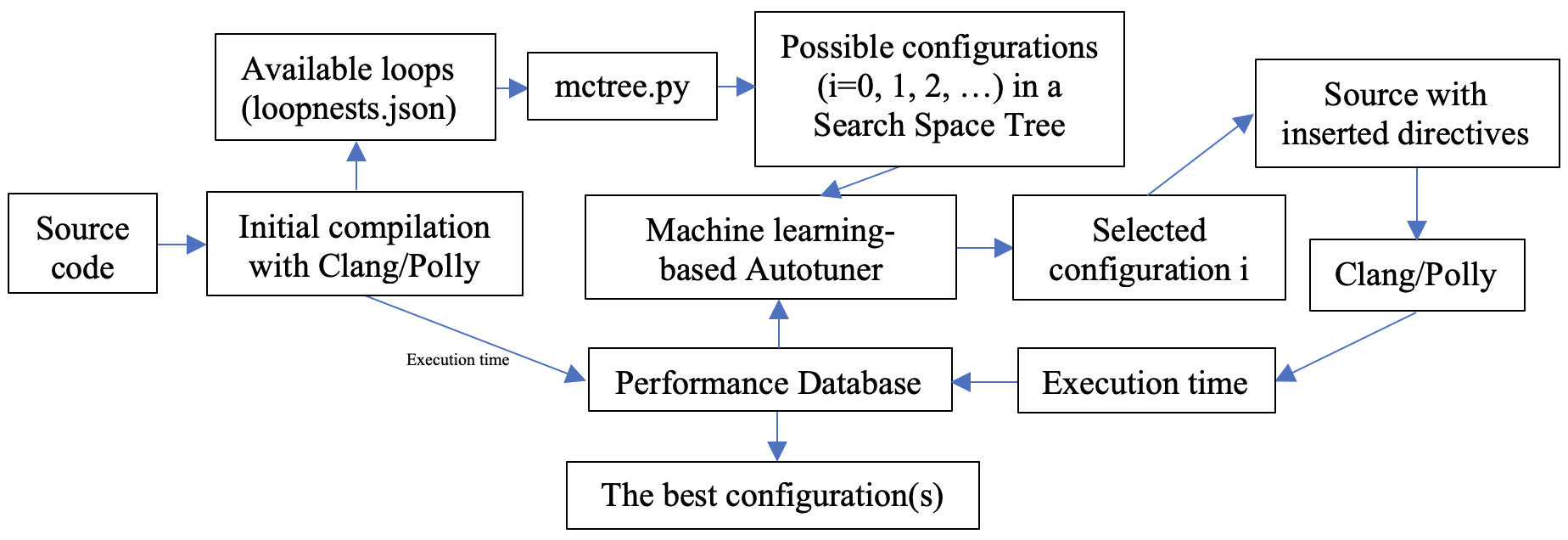}
 \caption{mctree autotuning framework}
\label{fig18}
\end{figure}

Recall that we presented the ytopt autotuning framework in Figure~\ref{fig2}, it requires a user to define the target parameters and parameterize the code to generate the code mold and build the parameter space. This is not convenient. In order to autotune a loop without a user's knowledge, we demonstrated a simple implementation mctree to show the viability of the autotuning search space for loop transformations that has the straightforward representation as either a tree or a directed acyclic graph \cite{MCTree, KF20}. We used the PolyBench benchmarks: gemm, syr2k, and covariance to evaluate the framework. Figure~\ref{fig18} shows the mctree autotuning framework, which consists of three components: the compiler for finding loop nests and applying transformations, the search space generator for deriving new configurations, and the autotuner for choosing configurations and executing commands. 

The source code without any loop transformations serves as the root node from which all other configurations are derived by adding loop transformations. As a result, the loop transformation search space has the structure of a tree by stacking up more and more loop transformations using a simple Python implementation mctree.py\cite{MCTree} based on available loop nests. Three different transformations in the search tree generator are implemented: tiling, loop interchange, and thread-parallelization. Internally, the loop nest is represented by an object tree; each object represents a loop and stores its unique name. 
After applying a transformation, the transformed loop objects are replaced with new ones representing the structure of the nest after the transformation.
For instance, tiling $n$ loops removes those objects and reinserts twice as many in their place. Interchange reinserts the same numbers of loops.
Thread parallelization inserts a loop object that is marked as non-transformable, i.e. an already parallel loop is not considered for additional transformations.
Loops not affected by a transformation keep their identifiers.
A simple autotuner using this search space is also implemented in mctree.py. 
An evaluation based on a selection of loop nests is conducted so that its execution time and configuration are stored in the performance database for the reference of next evaluation. At the end, the best configurations with the smallest time are the output by processing the performance database.

The PolyBench benchmark Floyd-Warshall was difficult to autotune in the previous section because Polly uses heuristics to optimize the benchmark that make it run much slower. For mctree autotuning we apply the three compiler options that avoid the problem as well.
In order to autotune the benchmark, a user just need to specify the loop kernel name \textbf{kernel\_floyd\_warshall} in the file floyd-warshall.c in the command line as follows:

{\scriptsize
\begin{verbatim}
python ../mctree.py autotune --keep --timeout 100 "clang" floyd-warshall.c polybench.c -O3 -DLARGE_DATASET 
       -march=native -mllvm -polly-only-func=kernel_floyd_warshall -o floyd-warshall
\end{verbatim}
 }
 
where the timeout is set as 100 seconds for each evaluation; the compiler is clang; the dataset is set as LARGE\_DATASET; the option -polly-only-func is set as \textbf{kernel\_floyd\_warshall}; and the executable name is floyd-warshall. The 13 tile size choices are set as \{4, 8, 16, 20, 32, 50, 64, 80, 96, 100, 128, 256, 2048\} used in the previous section. Note that the default tile size choices are set as \{2, 4\} in mctree.py.

\begin{figure}
\centering
\begin{minipage}{.5\textwidth}
  \centering
   \includegraphics[width=.6\textwidth]{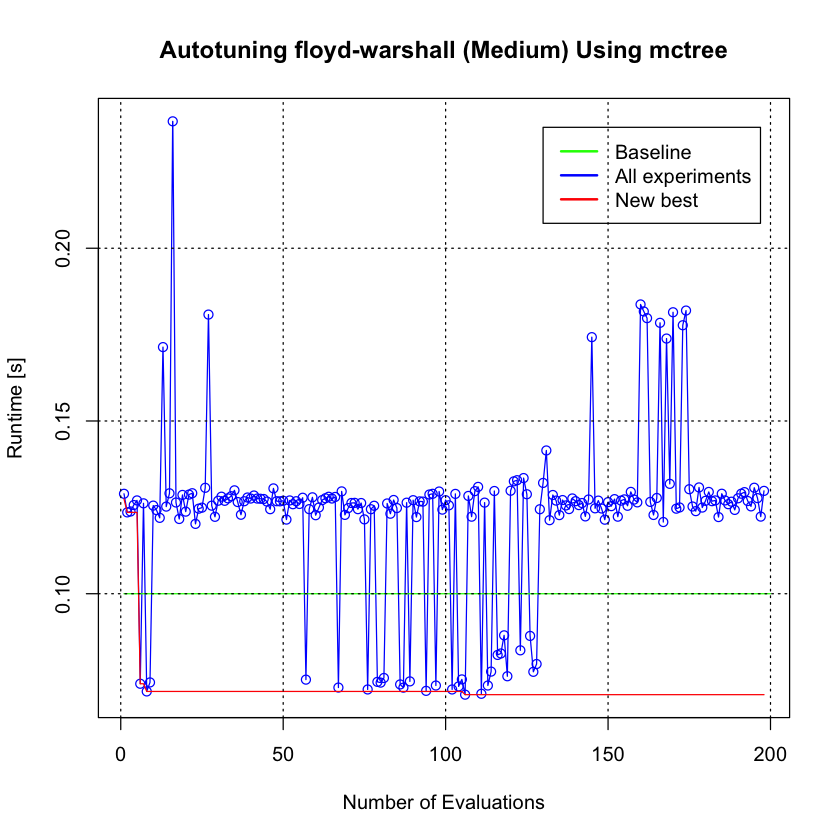}
\end{minipage}%
\begin{minipage}{.5\textwidth}
  \centering
 \includegraphics[width=.6\textwidth]{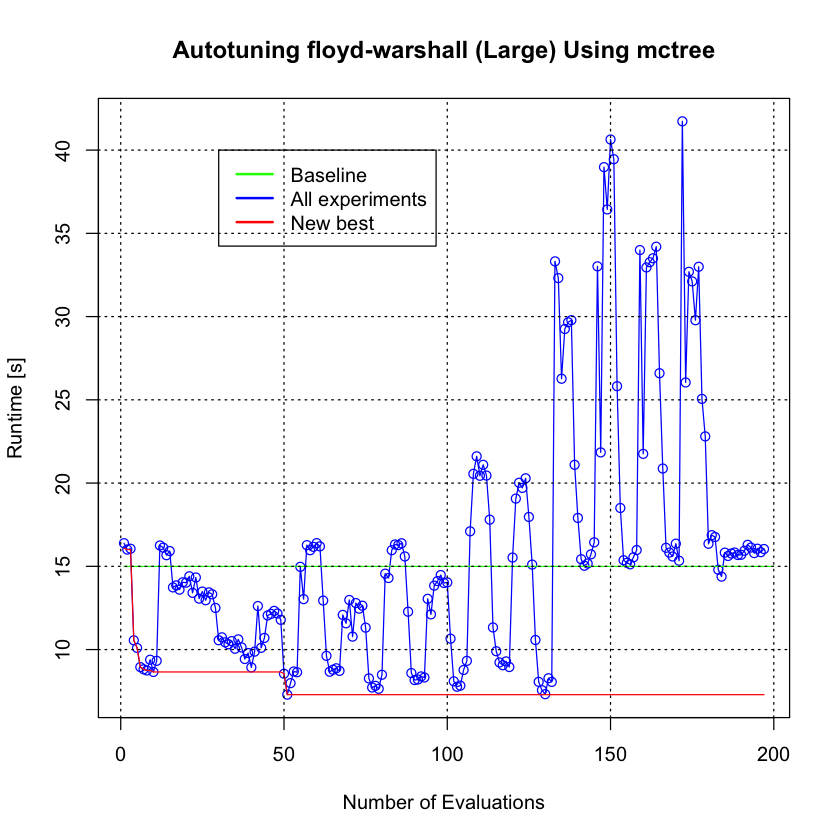}
\end{minipage}
 \caption{Autotuning Floyd-Warshall using mctree}
\label{fig13}
\end{figure}

Figure \ref{fig13} shows 200 valid evaluations for autotuning Floyd-Warshall with the medium and large datasets using mctree, where the green line stands for the smallest execution times: 0.076s for medium dataset in Table~\ref{tab7} and 14.995s for large dataset in Table~\ref{tab6} as the \textbf{baseline}; the blue line stands for all experiments for evaluation; and the red line stands for the new best runtime during the autotuning process. If the current experiment is faster than the previously fastest experiment,  the \textbf{new best} is lowered to the current runtime. At the end of the graphs, the red line shows the overall best runtime.

From Figure~\ref{fig13}, we observe that the overall best runtime using mctree is 0.0707s for the medium dataset with the configuration (\#pragma clang loop(loop2,loop3) tile sizes(8,64) floor\_ids(loop5362,loop5358) tile\_ids(loop5363,loop5359)) and 7.291s for the large dataset with the configuration (\#pragma clang loop(loop2,loop3) tile sizes(20,128) floor\_ids(loop5478,loop5470) tile\_ids(loop5479,loop5471)). This is a good improvement. This indicates that tiling two outer loops results in the best runtime. This explains why we could not achieve the better performance in the previous section because we applied loop tiling to all three loops for autotuning.

Note that, for the simple mctree autotuning framework, we did not implement any additional search pruning; instead we rely on Polly to reject any malformed transformation sequence. Also, detection of equivalent transformations through different paths has not been implemented. When we collected 200 valid evaluations, the mctree autotuning ran the total of 401 experiments indeed. We plan to continue to improve the search space generation by integration into the ytopt autotuning framework using a hybrid of Monte Carlo tree search exploration strategies and machine learning on vector spaces.

\section{Autotuning a Deep Learning Application}

In this section, we extend our ytopt autotuning framework to tune a deep learning application MNIST, which is implemented by using Keras \cite{KRS} and TensorFlow \cite{TFL} and multilayer perceptron (MLP). This deep learning application uses the MNIST database of handwritten digits (http://yann.lecun.com/exdb/mnist/). To achieve high test accuracy as a metric, we have to tune several parameters such as batch size, epochs, learning rate, dropout rate, optimizer, and so on.

\begin{figure}
\center
 \includegraphics[width=.6\textwidth]{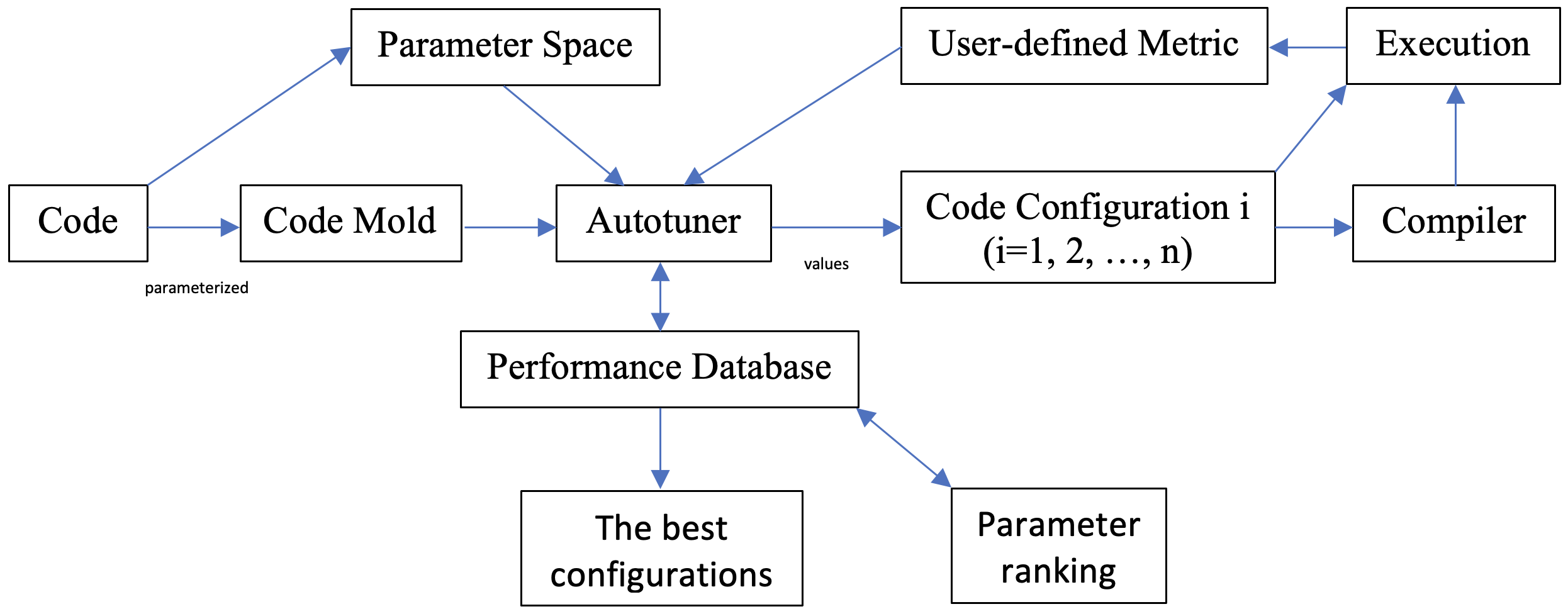}
 \caption{A general ytopt autotuning framework}
\label{fig11}
\end{figure}

Recall that we presented the ytopt autotuning framework in Figure \ref{fig2}, where the metric is the execution time. We leveraged the Bayesian optimization to explore the parameter space search to find high-quality configuration for the smallest runtime. In order to tune the hyperparameters for deep neural networks, we have to extend the autotuning framework to consider any user-defined metric shown in Figure \ref{fig11}. 
To that end, we define the metric as $1/accuracy$ so that we can use the framework to find high-performing configuration for the smallest $1/accuracy$. 


For MNIST, we define the following parameters: batch size (P0), epochs (P1), dropout rate (P2), and optimizer (P3). Based on these parameters, we define the following parameter space input\_space using ConfigSpace:

{\scriptsize
\begin{verbatim}
cs  CS.ConfigurationSpace(seed=1234)
P0= CSH.OrdinalHyperparameter(name='P0', sequence=['16','32','64','100','128','200','256','300','400','512'], 
       default_value='128')
P1= CSH.OrdinalHyperparameter(name='P1', sequence=['1','2','4','8','12','16','20','22','24','30'], 
       default_value='20')
P2= CSH.OrdinalHyperparameter(name='P2', sequence=['0.1', '0.15', '0.2', '0.25','0.4'], default_value='0.2')
P3= CSH.CategoricalHyperparameter(name='P3', choices=['rmsprop','adam','sgd','adamax','adadelta','adagrad',
       'nadam'], default_value='rmsprop')
cs.add_hyperparameters([P0, P1, P2, P3])
input_space = cs 
\end{verbatim}
 }
where the parameters P0, P1, P2, and P3 have the multiple choices. For simplicity, we set 10 choices for P0 and P1 and 5 choices for P2.  We set 7 choices for P3 because Keras supports 7 optimizers (https://keras.io/api/optimizers/). This parameter space consists of 10x10x5x7= 3,500 different configurations. We use the general autotuning framework to find out which configuration results in the highest test accuracy and use four ML methods---RF, ET, GBRT, and GP---to investigate which ML method generates the smallest 1/accuracy for which configuration in 200 evaluations.

\begin{figure}
\centering
\begin{minipage}{.5\textwidth}
  \centering
 \includegraphics[width=.6\textwidth]{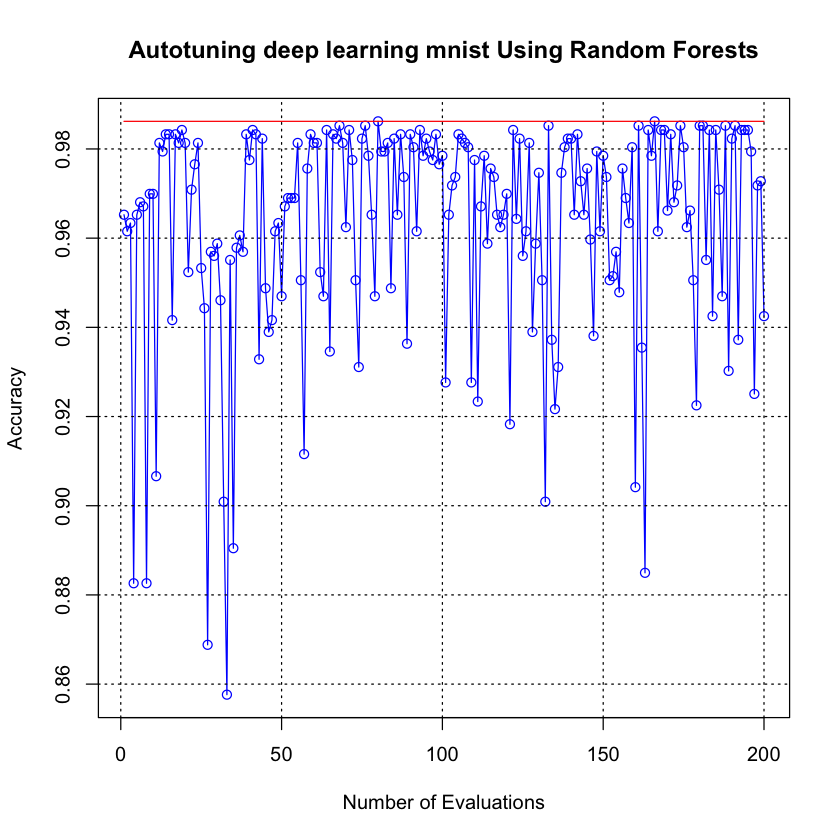}
 \caption{Autotuning mnist using RF in 200 evaluations}
\label{fig14}
\end{minipage}%
\begin{minipage}{.5\textwidth}
  \centering
 \includegraphics[width=.6\textwidth]{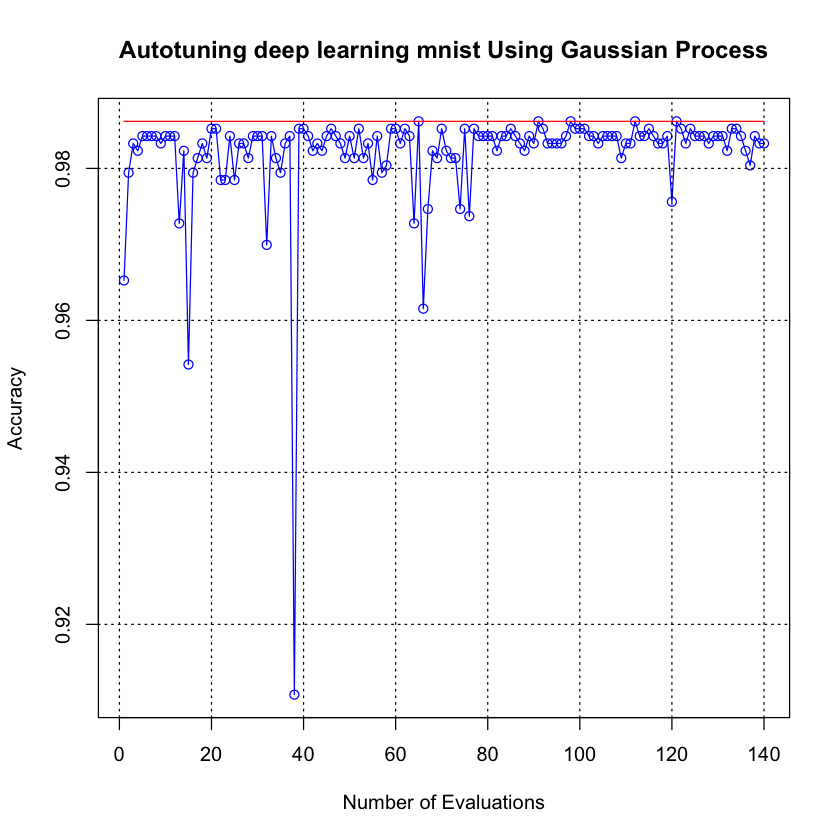}
 \caption{Autotuning mnist using GP in 200 evaluations}
\label{fig15}
\end{minipage}
\end{figure}

In Figure \ref{fig14}, RF results in the highest accuracy of 0.986 for the configurations (128, 24, 0.4, 'adadelta') at Evaluation 80 and (16, 22, 0.1, 'adadelta') at Evaluation 166 of 200 evaluations. Note that the configuration (128, 24, 0.4, 'adadelta') stands for the combination of batch size of 128, epochs of 24, dropout rate of 0.4, and the optimizer adadelta. The blue line is for all evaluations; the red line is the highest test accuracy. For the deep learning application, batch steps per epoch is the total number of samples divided by the batch size. Increasing the batch size means decreasing the batch steps per epoch which results in the smaller training time. The tradeoff based on the same accuracy is to choose the configuration (128, 24, 0.4, 'adadelta') with the smaller training time.

In Figure \ref{fig15}, GP results in the same highest accuracy of 0.986 for the configurations (128, 22, 0.2, 'adadelta') at Evaluation 65, (128, 24, 0.4, 'adadelta') at Evaluation 91, (128, 24, 0.4, 'adamax') at Evaluation 98, (32, 24, 0.2, 'adamax') at Evaluation 112, (16, 30, 0.2, 'adadelta') at Evaluation 121 of 140 evaluations. As we mentioned before, GP uses random sampling to generate the parameter configurations for performance evaluation. In this case, 140 of 200 evaluations were conducted, the other 60 evaluations are skipped because of the replicated evaluations. The configuration (128, 22, 0.2, 'adadelta') results in the smallest runtime as well.

\begin{figure}
\centering
\begin{minipage}{.5\textwidth}
  \centering
 \includegraphics[width=.6\textwidth]{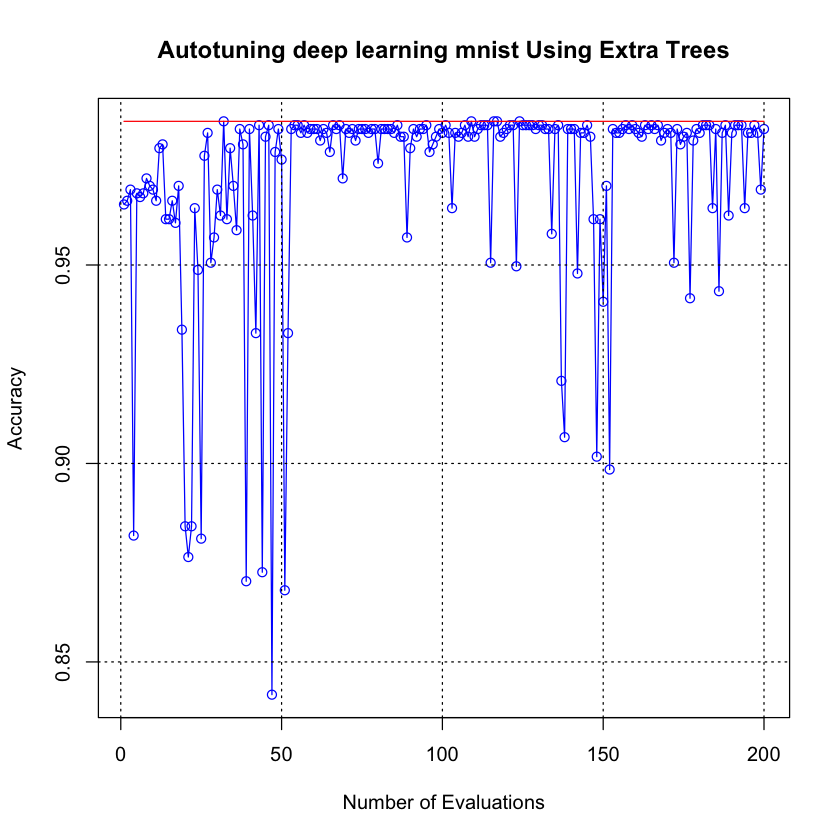}
 \caption{Autotuning mnist using ET in 200 evaluations}
\label{fig16}
\end{minipage}%
\begin{minipage}{.5\textwidth}
  \centering
 \includegraphics[width=.6\textwidth]{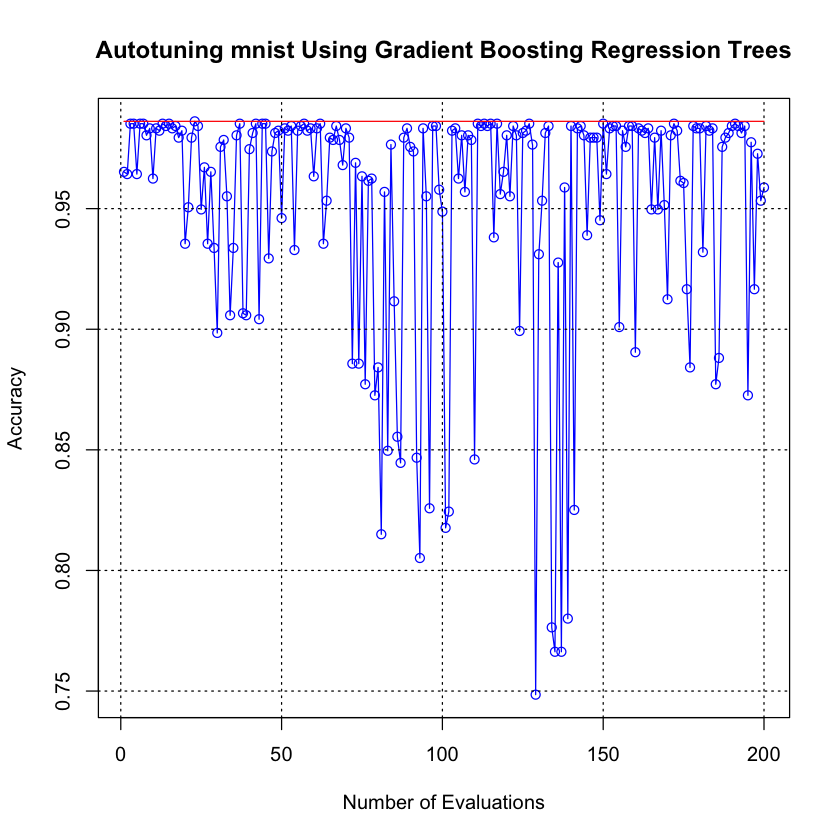}
 \caption{Autotuning mnist using GBRT in 200 evaluations}
\label{fig17}
\end{minipage}
\end{figure}

In Figure \ref{fig16}, ET results in the same highest accuracy of 0.986 for  the configurations (100, 12, 0.1, 'adamax') at Evaluation 32, (16, 12, 0.2, 'adadelta') at Evaluation 96, (128, 16, 0.2, 'adadelta') at Evaluation 109, (128, 22, 0.2, 'adadelta') at Evaluation 116, (16, 20, 0.2, 'adadelta') at Evaluation 117 of 200 evaluations. We observe that the configuration (100, 12, 0.1, 'adamax') results in the smallest runtime as well.
In Figure \ref{fig17}, GBRT results in the highest accuracy of 0.986 for the configuration (100, 12, 0.2, 'adadelta') at Evaluation 23 of 200 evaluations. 

Overall, we observe that using all four ML methods results in the same highest test accuracy based on the fixed parameter space. GP results in the smallest overhead. For the MNIST, using the optimizers adadelta or adamax led to the highest accuracy of 0.986 for the configurations (100, 12, 0.2, 'adadelta') and (100, 12, 0.1, 'adamax') with the smallest runtime. So the general ytopt autotuning framework can be applied to other deep learning applications.

\section{Related Work}

A large amount of literature on autotuning exists. Balaprakash et al.\cite{BD18} surveyed the state of the practice in incorporating autotuned code into HPC applications; the authors highlighted insights from prior work and identified the challenges in advancing autotuning into wider and long-term use. Traditional autotuning methods are built on heuristics that derive from automatically tuned BLAS libraries \cite{ATLAS}, experience \cite{TC02, CH06, GO10} and model-based methods \cite{Chen05, TH11, BG13, FE17}. At the compiler level \cite{AK18}, machine-learning-based methods are used for automatic tuning of the iterative compilation process \cite{OP17} and tuning of compiler-generated code \cite{TC09, MS14}. Recent work on autotuning of OpenMP code has gone beyond loop schedules to look at parallel tasks and function inlining \cite{SJ19, KC14, MA11, Rose09}.  Some recent work has used machine learning and sophisticated statistical learning methods to reduce the overhead of autotuning \cite{RB16, MA17, TN18, BQ19}.

There are two critical requirements of autotuning: 1) expression of a search space of implementations/configurations, and 2) efficient navigation of the search space for the optimal configuration. To address these two requirements, a number of autotuning frameworks have been developed that interface with application code, libraries and compilers to generate code variants and measure their performance\cite{TC02,TH11,HN09,AK14, NR15,ZG15, RH17,CLTune,PG19,KernelTuner,HG20,YTO,WK20}. They present the expression of a collection of parameters to be tuned and their corresponding possible values and generate possible configurations which may or may not be valid for evaluation. 

There exist two kinds of expressions of search space: vector space and tree space. 
Most of autotuning frameworks present the search space in a vector space, that is, a fixed number of parameter knobs, such as OpenTuner \cite{AK14}, CLTune \cite{CLTune}, HalideTuner \cite{ZG15}, Orio \cite{HN09}, KernelTuner \cite{KernelTuner}, AFT \cite{RH17, RS21}, ytopt \cite{YTO, WK20}, etc. The successor of HalideTuner \cite{AM19} uses tree search to avoid the limitation of a vector search space, but uses Beam search to explore the space. ProTuner~\cite{HG20} further improved Halide schedule autotuning by replacing Beam search with Monte Carlo tree search. We also demonstrated the viability of the autotuning search space for loop transformations that has the straightforward representation as either a tree or a directed acyclic graph using mctree \cite{MCTree, KF20}.
The loop autotuner in Telamon also uses Monte-Carlo Tree Search\cite{Telamon}. In the tradition of Halide every level needs assigned a strategy and a schedule where not all loops have an assigned strategy is considered incomplete. Whereas in our approach, every loop is considered sequential until we add a pragma.

We classify the existing autotuning frameworks into four categories:1) enumerate all possible parameter configurations, reject invalid ones, and evaluate the valid ones \cite{KF20}; 2) enumerate only valid configurations~\cite{RH17, RS21};  3) sample from the set of possible configurations, and reject invalid ones~\cite{HN09, NR15, SJ19} during the search; 4) sample only valid configurations and search over them \cite{WK20}. Our ytopt autotuning framework belongs to Category 4 which overcomes the ineffectiveness of Category 3 by generating valid samples and addresses the limitations of Categories 1 and 2, where enumerating all possible configurations can be computationally expensive for large number of parameters. 


\section{Summary and Future Work}

We developed the ML-based ytopt autotuning framework, and applied the newly developed Clang loop optimization pragmas to six complex PolyBench benchmarks to optimize them. We defined the parameters for these pragmas and then used the autotuning framework to optimize the pragma parameters to improve their performance. We evaluated the effectiveness of four different supervised ML methods used as the surrogate model within Bayesian optimization for each benchmark. The autotuning outperformed the other compiling methods to provide the smallest execution time for the benchmarks syr2k, 3mm, heat-3d, lu, and covariance with both large datasets. An exception is the Floyd-Warshall benchmark because Polly uses heuristics to optimize the benchmark to make it run much slower. To cope with this situation, we provide three compiler option solutions to improve the performance. Then we presented loop autotuning without a user's knowledge using a simple mctree autotuning framework to further improve the performance of the Floyd-Warshall benchmark. We also extended the ytopt autotuning framework to tune the deep learning application MNIST. This ytopt autotuning framework is open source and available from the link in \cite{YTO, WK20}, and the simple mctree autotuning framework is also open source and available from the link \cite{MCTree, KF20}. 

For future work, we will easily extend the current ytopt autotuning framework to support various HPC applications because of our symbol representations for the pragmas and related parameters. In \cite{SP18}, autotuning OpenMP codes was investigated for energy efficient HPC systems. In \cite{WM20}, an end-to-end autotuning framework in HPC PowerStack was proposed to tune the power and energy ecosystem. We will extend this ytopt autotuning framework to consider power consumption and energy consumption. When we consider power or energy as the optimal solution, this may change how to do the parameter space search based on the new metric. These parameters can be extended to include application parameters, system environment parameters such as setting number of threads, thread scheduling and affinity, JIT-enabled parameters, power-capping size, and loop transformation parameters, and so on. Future work will also focus on improving loop autotuning without a user's knowledge by integration into the ytopt autotuning framework using a hybrid of Monte Carlo tree search exploration strategies and machine learning on vector spaces.

\section{Acknowledgments}
This work was supported in part by LDRD funding from Argonne National Laboratory, provided by the Director, Office of Science, of the U.S. Department of Energy (DoE) under contract DE-AC02-06CH11357, in part by DoE ECP PROTEAS-TUNE, and in part by NSF grant CCF-1801856.

\bibliography{wileyNJD-AMA}%

\end{document}